\pdfoutput=1
\documentclass[letterpaper, 10 pt, conference]{ieeeconf}  

\IEEEoverridecommandlockouts                              
\usepackage{graphicx} 
\usepackage{multicol}
\usepackage{amsmath} 
\usepackage{amssymb}  
\usepackage{array}
\usepackage{subcaption}

\title{\LARGE \bf
Simultaneous Localization and Layout Model Selection \newline in Manhattan Worlds
}

\author{Armon Shariati$^{1}$, Bernd Pfrommer$^{1}$, Camillo J. Taylor$^{1}$
\thanks{$^{1}$GRASP Lab, University of Pennsylvania, Philadelphia PA 19104. \texttt{\{ armon, pfrommer, cjtaylor \} @seas.upenn.edu}}%
}

\begin{document}

\maketitle
\thispagestyle{empty}
\pagestyle{empty}

\begin{abstract}
In this paper, we will demonstrate how Manhattan structure can be exploited to transform the Simultaneous Localization and Mapping (SLAM) problem, which is typically solved by a nonlinear optimization over feature positions, into a model selection problem solved by a convex optimization over higher order layout structures, namely walls, floors, and ceilings. Furthermore, we show how our novel formulation leads to an optimization procedure that automatically performs data association and loop closure and which ultimately produces the simplest model of the environment that is consistent with the available measurements. We verify our method on real world data sets collected with various sensing modalities. 
\end{abstract}

\section{Introduction}

This paper describes a novel approach to Simultaneous Localization and Mapping (SLAM) that leverages the Manhattan structure of the environment and reformulates the reconstruction task in terms of a model selection problem. Importantly, the key subproblem of establishing long range correspondences and loop closures is incorporated into the reconstruction procedure and solved automatically as part of the optimization procedure.

While there exist several SLAM approaches that try to incorporate the rectilinear structure of indoor man-made environments within their models by tracking more semantically meaningful features such as lines and planes \cite{Hsiao2018,Pumarola2017}, to our knowledge, we are the first to frame the mapping aspect of the problem entirely as one of model selection. Our ideal model is one that would resemble an architect's blueprint which outlines the location of all large static \emph{layout structures}, namely walls, floors, and ceilings. Such a generative model, even partially complete, could not only enable a robot to track it's position and orientation within the environment with greater precision, but could also serve as a strong prior for occupancy inference, as well as object detection and completion. We demonstrate on real data that our novel formulation, based on the principle of Occam's razor (i.e. select the simplest model of those that best describe the data), produces an optimized trajectory and a compact map representation, jointly.

\section{Related Work}

This work aims to simultaneously address several problems which collectively intersect a few broader research topics in robotics and computer vision. At the heart of our system is an optimization over a series of robot poses and landmarks, which clearly places it squarely in the SLAM domain. On the other hand, our contributions to automated layout-model selection extends its relevance to loop closure and automated floor plan generation. 

\subsection{Simultaneous Localization and Mapping}

The SLAM problem has been a central concern in robotics research for well over two decades. While space constraints prohibit a comprehensive review, the dichotomy highlighted in \cite{Li2018} between non-structural and structural SLAM systems is quite useful when considering where the contributions of this paper best fit. 

The more popular and mature non-structural SLAM systems such as \cite{Li2017,Mur-Artal2015a,Pumarola2017} prioritize generalizability across environments at the cost of accuracy as they refrain from introducing any hard constraints between landmarks and poses, and among landmarks themselves. Meanwhile, the recently popular structural SLAM systems such as \cite{Camposeco2015,Hsiao2018,Straub2017,Zhou2015} leverage structural cues in order to provide geometric constraints, which can be used to improve accuracy at the cost of universality. Our present work falls in the latter category.

The works most closely related to our own among this body of literature would be that of Kim \emph{et al.} \cite{Kim2018,Kim}. The first tries to use structural regularities to reduce drift by decoupling estimates for rotation and translation, much like we do, while the second involves a nearly identical initial approach and output, and focuses on the computational limitations of previous approaches using a Bayesian filtering based technique. However, neither algorithm seeks to address the issues surrounding data association and automatic model selection. Similarly, the work of \cite{Hsiao2018} also bears some resemblance to our own in terms of an approach and the tools they use, yet they too overlook the challenge of compact map synthesis and loop closure.  

Semantic SLAM \cite{Bowman2017a} is a relatively new body of work which shares many of our own objectives in a more general context. However, a key component of our approach is the assumption of underlying rectilinear structure which we explicitly exploit to improve the quality of our reconstruction. 

\subsection{Loop Closure}

As the results in this paper are also relevant to visual loop closure, it is worth visiting at least a few approaches to the problem thus far. However, a more in depth review of the subject may be found in \cite{Lowry2016}. 


What distinguishes our work here from many of the other key-frame based SLAM systems that also perform loop detection and closure, such as \cite{Mur-Artal2015a}, is that our system is able to solve the loop detection problem and the belief-update problem in a single optimization framework instead of relying on separate modules. 
While this combinatorial optimization problem is NP-hard we are able to reformulate the search as a convex optimization problem which makes this approach feasible.

Latif \emph{et al.} \cite{Latif2017} also observed that sparsity can be leveraged in the context of loop closure. However, while their approach is effectively a search through a set of correspondences to find one sparse set of basis vectors for reconstruction, we invert the problem by trying to instead maximize the total number of correspondences, which allows us to solve the more general problem of data association across temporal frames in addition to loop closure. 

\subsection{Automated Floor Plan Generation}

Constructing a meaningful representation of an indoor environment is at the heart of automated floor plan generation \cite{Okorn2010}. While we may share many of the motivations behind the works within this body of literature, our problem domain is that of robotic exploration. As a result, the complexity of many of the models used in these approaches are still greater than is necessary, and furthermore, entirely overlook the localization aspect of the SLAM problem.

Among the more complex and expressive models, are those described in \cite{Mura2016} and \cite{Ochmann2016}, both of which lift the Manhattan World assumption in order to output a set of watertight polyhedra representing the boundaries of rooms in the environment. These techniques also focus on volumetric labeling of the space.

While both \cite{Liu2018a} and \cite{Angladon2018} present solutions for use on mobile platforms, Liu \emph{et al.} take a learning approach to the problem, whereas the authors of \cite{Angladon2018} leverage the potential of user intervention.

Another approach, relying on image panoramas alone can be found in \cite{Cabral2014}, which formulates the floor plan reconstruction as a shortest path problem.

\section{Technical Approach}

\begin{figure}
	\centering
	\includegraphics[width=0.6\columnwidth]{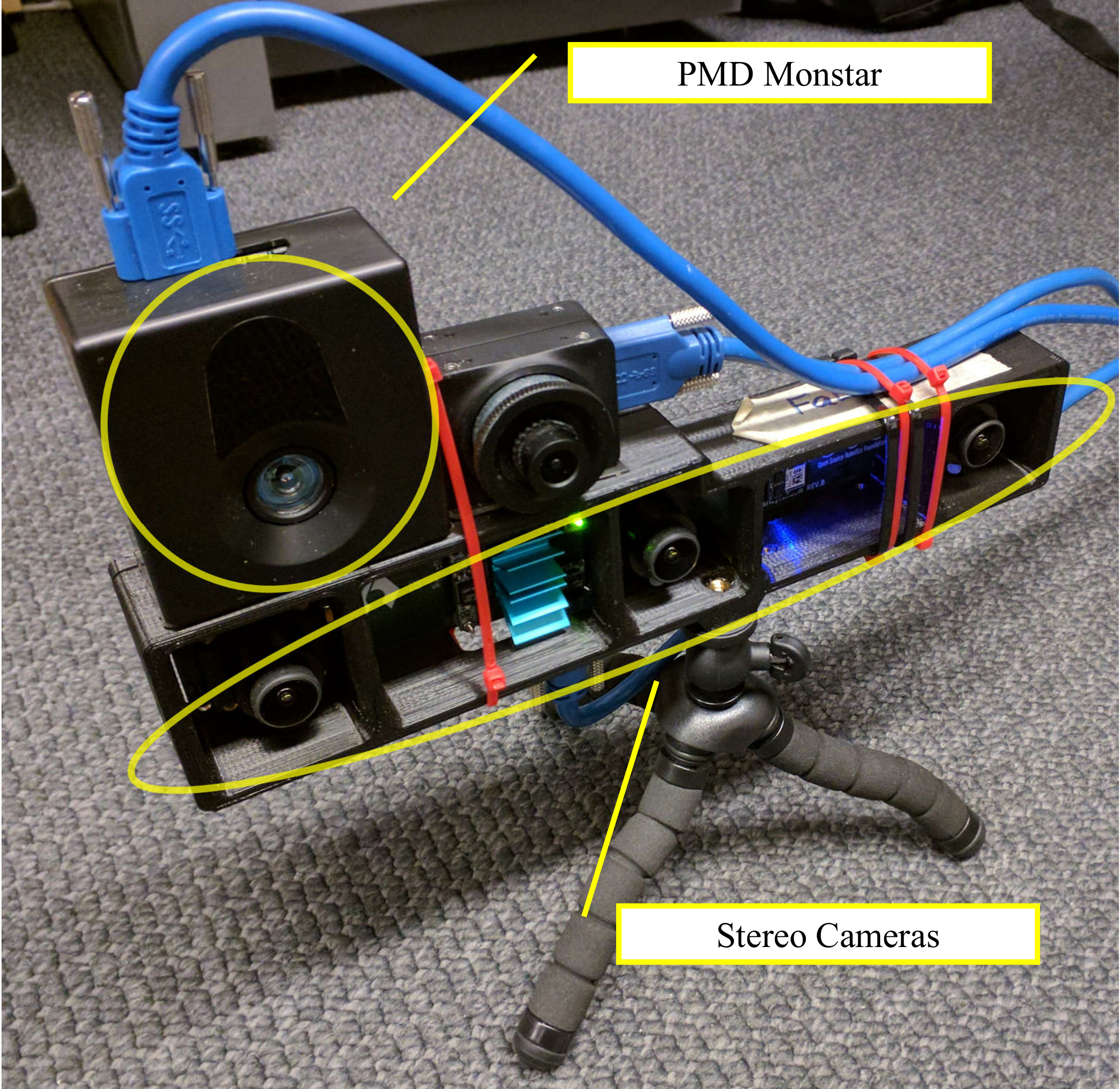}
    \caption{Sensor rig used to acquire data. Annotated in yellow are the PMD Monstar depth sensor and a custom stereo pair. The camera to the right of the Monstar and the center stereo camera are not used. }
    \label{fig:falcam}
\end{figure}

In this section we describe the main features of our structural analysis procedure. To ground our subsequent discussion we begin with a brief description of the input data and the hardware systems used to acquire it in our subsequent experiments. Figure \ref{fig:falcam} shows an example of one of our sensor rigs. Despite slight differences among sensor configurations, every rig features a stereo pair of cameras, hardware synchronized to an inertial measurement unit (IMU), as well as a depth sensor that captures low-resolution depth images up to a range of about 6 meters. The data from the stereo cameras and IMU are used to drive a stereo based MSCKF Visual-Inertial Odometery (VIO) system \cite{Sun2017a}.  Further details surrounding each sensor are provided in Section \ref{sec:exp}.

The end result is that our sensor suites provide the analysis algorithm with a set of depth maps along with initial estimates for the relative motion of the sensor rig over time and an estimate for the direction of the gravity vector in each frame.


We note that similar datasets could be acquired using other means. One could acquire depth maps using a LIDAR sensor like the Velodyne puck or from a passive stereo system. Similarly pose information could be derived from monocular Visual-Inertial Odometry or from wheel encoders on a moving platform. The proposed analysis would still be applicable in all of these cases.

The first stage in the analysis involves processing each frame in the depth map separately to extract salient axis-aligned planar fragments of layout structure, called \emph{layout segments}. The first step in this process involves projecting the depth points into the plane defined by the measured gravity vector and then rotating the resulting 2D point set in one degree increments to find a yaw orientation that minimizes the entropy of the resulting point distribution. This algorithm is described in a number of previous works including \cite{Bazin2013,Taneja2015}, and \cite{Cowley2011} where it is referred to as an entropy compass. Upon completion, this procedure recovers the orientation of the frame with respect to the prevailing Manhattan structure. We note that one could also use other means for Manhattan frame estimation such as \cite{Straub2015,Joo2016}.

\begin{figure}
	\centering
    \includegraphics[width=0.7\columnwidth]{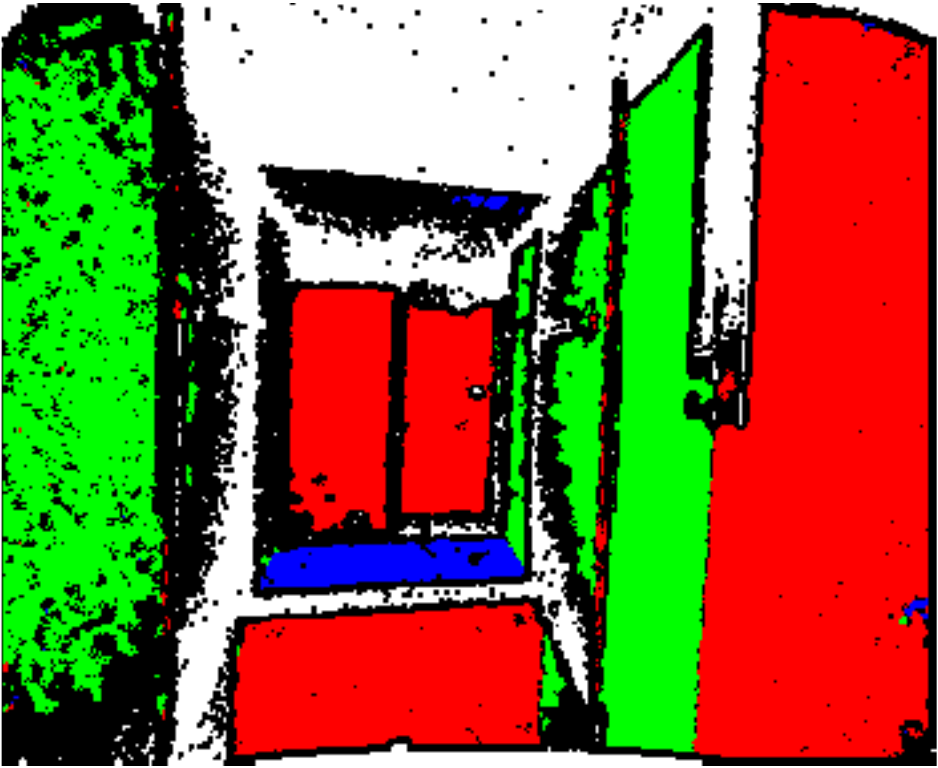}
    \caption{Depth map broken into salient surfaces. Red, green, and blue pixels represent $x$, $y$, and $z$-axis alignment. A pixel $p$ is assigned to the major axis which maximizes the number of pixels in its $k \times k$ neighborhood that would reside on the plane centered at $p$ with the given major axis orientation. If no axis can be assigned with sufficient confidence or no depth information is recorded at $p$, it is colored black and white respectively. }
    \label{fig:assignment}
\end{figure}

Once this has been done, the system labels each pixel in the depth map according to the axis alignment of the surrounding $k \times k$ patch, and then groups them together using a connected components procedure as shown in Figure \ref{fig:assignment}. Finally, using an inverse perspective projection, each cluster of pixels is projected into the aligned sensor frame as a point cloud where we fit an orientation-constrained planar model using RANSAC. Planar segments with insufficient extent are discarded to favor the detection of dominant structures.

In addition to the entropy compass procedure which is applied to each frame individually, the system has an estimate for the relative orientation between each frame derived from the visual-inertial odometry system. These two sources of information are fused to provide a final estimate of the orientation of each frame in the sequence. The relative yaw estimates from the VIO system are used to constrain the range of angles considered in the entropy compass phase and to provide orientation estimates during periods where no axis-aligned surfaces are visible.

\begin{figure}[t]
	\centering
	\includegraphics[width=\columnwidth]{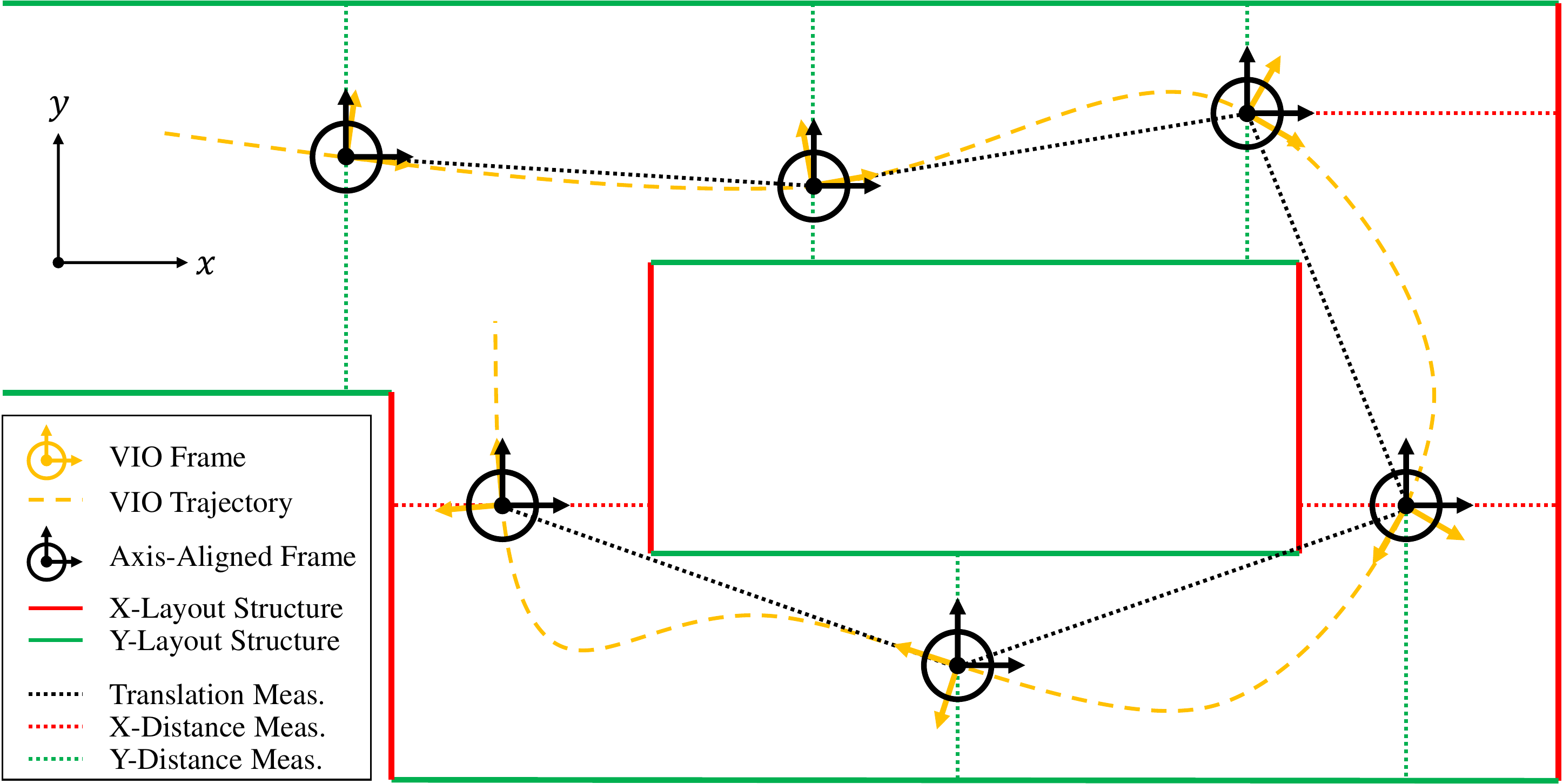}
	\caption{A 2-dimensional geometric representation of our model which illustrates a sensor moving through a Manhattan environment making periodic range measurements to various layout structures. Solid lines correspond to layout structures, while dotted lines correspond to measurements. Each distance measurement to a particular layout structure corresponds to the distance computed to the visible layout segment within the depth map captured at that frame. }
	\label{fig:geometric}
\end{figure}

The end result of the procedure is described in Figure \ref{fig:geometric}, which shows a top down view of a set of camera frames. Each frame is associated with two coordinate frames of reference, one which indicates the actual orientation of the sensor head and the other indicating an axis-aligned frame derived from the entropy analysis. Each layout measurement, denoting an estimate for the minimum distance between the camera frame and the corresponding layout segment observed at that frame, is depicted by a red or green dotted line.

\begin{figure}[t]
	\centering
	\includegraphics[width=\columnwidth]{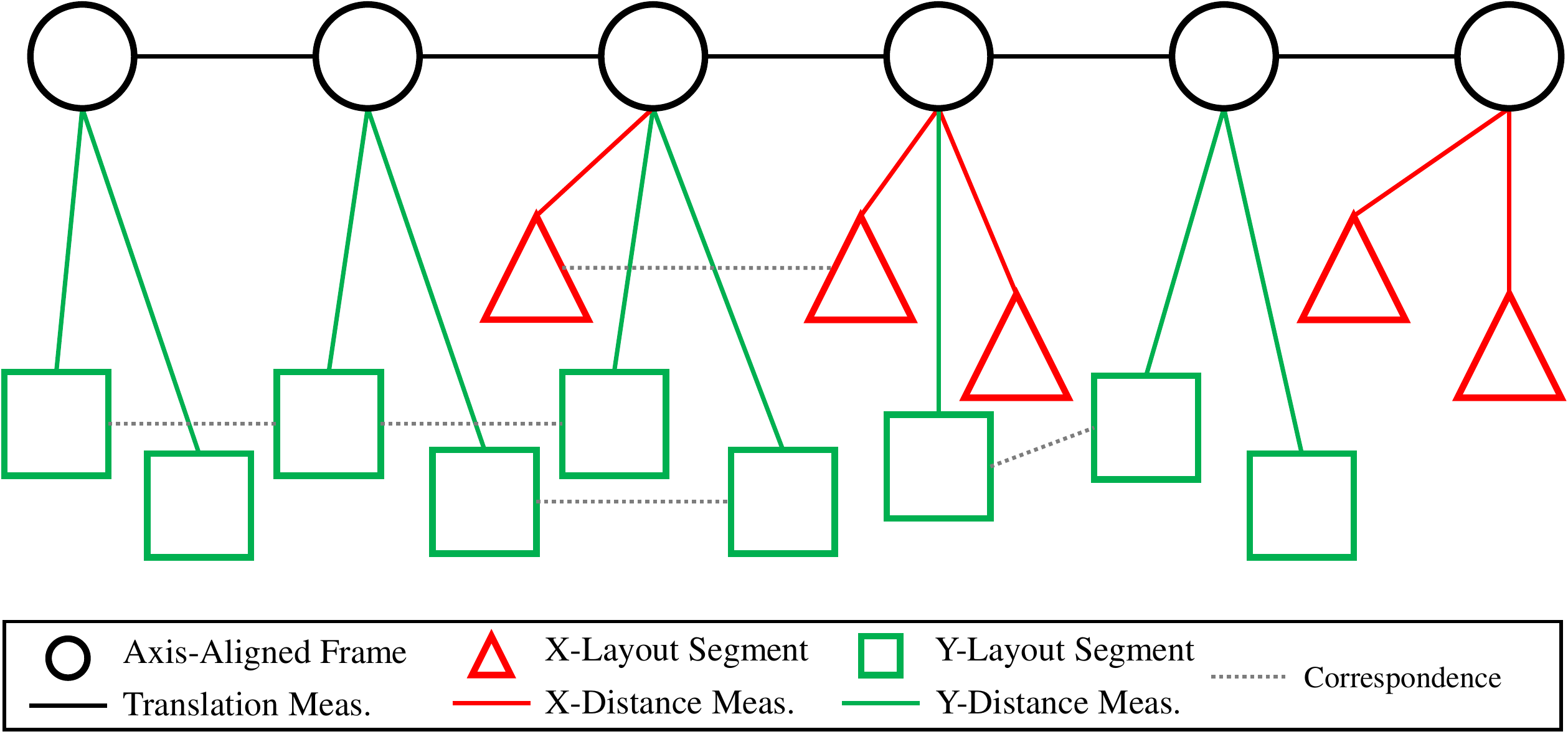}
	\caption{A functional representation of our model, in two dimensions, as a factor-graph. Circles correspond to robot locations, while triangles and squares correspond to x and y aligned layout segments respectively. Solid lines correspond to measurement factors derived from the VIO, entropy analysis, and depth map processing. We extend the traditional factor-graph formulation by including binary correspondence edges, represented by dotted lines. Initially generated by a temporal analysis, the set of hypothetical correspondence edges is also augmented by a user defined heuristic. Our sparse optimization procedure ultimately determines which of these constraints to enforce and discard. }
	\label{fig:functional}
\end{figure}

This system of measurements can be abstracted into the factor-graph \cite{Kschischang2001} shown in Figure \ref{fig:functional}. Here the circular nodes on top correspond to axis-aligned frame positions while the triangular and rectangular nodes on the bottom correspond to layout segments. The links between frames correspond to the estimates for inter-frame motion while the links between the frames and the layout segments correspond to the distance measurements described in Figure \ref{fig:geometric}. Expressing each estimate of inter-frame translation, provided by the VIO subsystem, with respect to the previous estimate of the Manhattan frame, provided by the orientation estimation procedure, yields an axis-aligned and -- in principle -- drift-reduced trajectory.

In the sequel we will use the following notation to describe the elements of the model shown in Figure \ref{fig:functional}. Let $\mathbf{p}_i \in \mathbb{R}^3$ denote the position of frame $i$ in the axis-aligned trajectory while $R_i \in \mathbf{SO}(3)$ denotes the orientation of the axis-aligned frame with respect to the corresponding sensor frame. Each of the layout segments that we observe will ultimately be associated with a structural supporting \emph{layout plane}, which is modeled as an axis-aligned surface with infinite extent. Each such layout plane will be modeled with a single parameter. More specifically we will let $m_j^x$ denote the $x$ coordinate of an infinite layout plane with index $j$ that is perpendicular to the $x$-axis of the model, similarly $m_k^y$ denotes the $y$ coordinate of a $y$ aligned layout plane with index $k$ and $m_l^z$ denotes the $z$ coordinate of a $z$ aligned layout plane with index $l$.

Correspondences between layout segments are denoted by the dotted lines in Figure \ref{fig:functional}. These correspondences amount to asserting that two extracted segments lie on the same axis-aligned layout plane. Note that these correspondences would typically link layout segments extracted in different frames but could also link two segments extracted in the same frame. At this stage of the analysis procedure a simple temporal analysis procedure is used to establish correspondences between segments seen in one frame and segments seen in the subsequent frame that have sufficient overlap. This initial set of correspondences will be augmented with longer range correspondences that are automatically discovered in a subsequent step of the process.

We will let the vector $\mathbf{t}_i \in \mathbb{R}^3$ denote the estimate for the translation between subsequent axis-aligned frames in the sequence that is derived from the visual odometry system and corrected by the orientation estimation procedure. That is $\mathbf{t}_i$ denotes an estimate for the quantity $\mathbf{p}_{i+1} - \mathbf{p}_i$.

We will let $\xi$ denote a vector formed by stacking the free parameters of our model, that is $\mathbf{p}_i$, $m_j^x$, $m_k^y$, and $m_l^z$, for all $i$, $j$, $k$, and $l$. Note that we assume that the camera orientations that align the frames with the Manhattan model, $R_i$, have been estimated using the entropy compass procedure described previously.

In this case the measurement system takes on a particularly simple linear form. Namely for each measurement from a frame to an $x$-aligned layout segment we have an equation of the form
\begin{equation}
	m_j^x - p_i^x = d_{ij}
    \label{eq:distx}
\end{equation}
where $m_j^x$ denotes the x coordinate associated with the $x$-aligned layout plane associated with the layout segment, $p_i^x$ denotes the $x$ coordinate of the position of frame $i$, and $d_{ij}$ denotes the measured offset between the layout segment and the camera as depicted in Figure \ref{fig:geometric}. Note $d_{ij}$ can be signed depending upon where the frame is relative to the layout segment. 

For layout segments aligned with the $y$ axis and $z$ axis we would have exactly analogous equations
\begin{equation}
	m_k^y - p_i^y = d_{ik}
    \label{eq:disty}
\end{equation}
\begin{equation}
	m_l^z - p_i^z = d_{il}
    \label{eq:distz}
\end{equation}

As previously discussed, the measurements of interframe motion derived from the VIO system and entropy analysis can be modeled as follows
\begin{equation}
	\mathbf{p}_{i+1} - \mathbf{p}_i = \mathbf{t}_i
    \label{eq:trans}
\end{equation}

Given this system of measurements the task of finding the optimal estimate for the structure of the scene and the trajectory of the sensor based on the factor-graph simply amounts to solving a sparse linear system $A \xi = b$ in a least squares sense. 
\begin{equation}
	\begin{aligned}
		& \underset{\xi}{\text{minimize}} && \|A \xi - \mathbf{b} \|_2 \\
	\end{aligned}
    \label{eq:linearopt}
\end{equation}

This is simply the system formed by stacking the measurement equations, namely Equations \ref{eq:distx}, \ref{eq:disty}, \ref{eq:distz}, and \ref{eq:trans}, into a single sparse system. The vector $\mathbf{b}$ aggregates the right hand sides of the equations including the distance measurements, $d_{ij}$, $d_{ik}$, $d_{il}$, and the translation estimates $\mathbf{t}_i$. This sparse system can be solved extremely efficiently even for relatively large systems of measurements.

\begin{figure}
    \centering
    \includegraphics[width=0.7\columnwidth,clip,trim={0.5in 0.75in 0.75in 0.75in}]{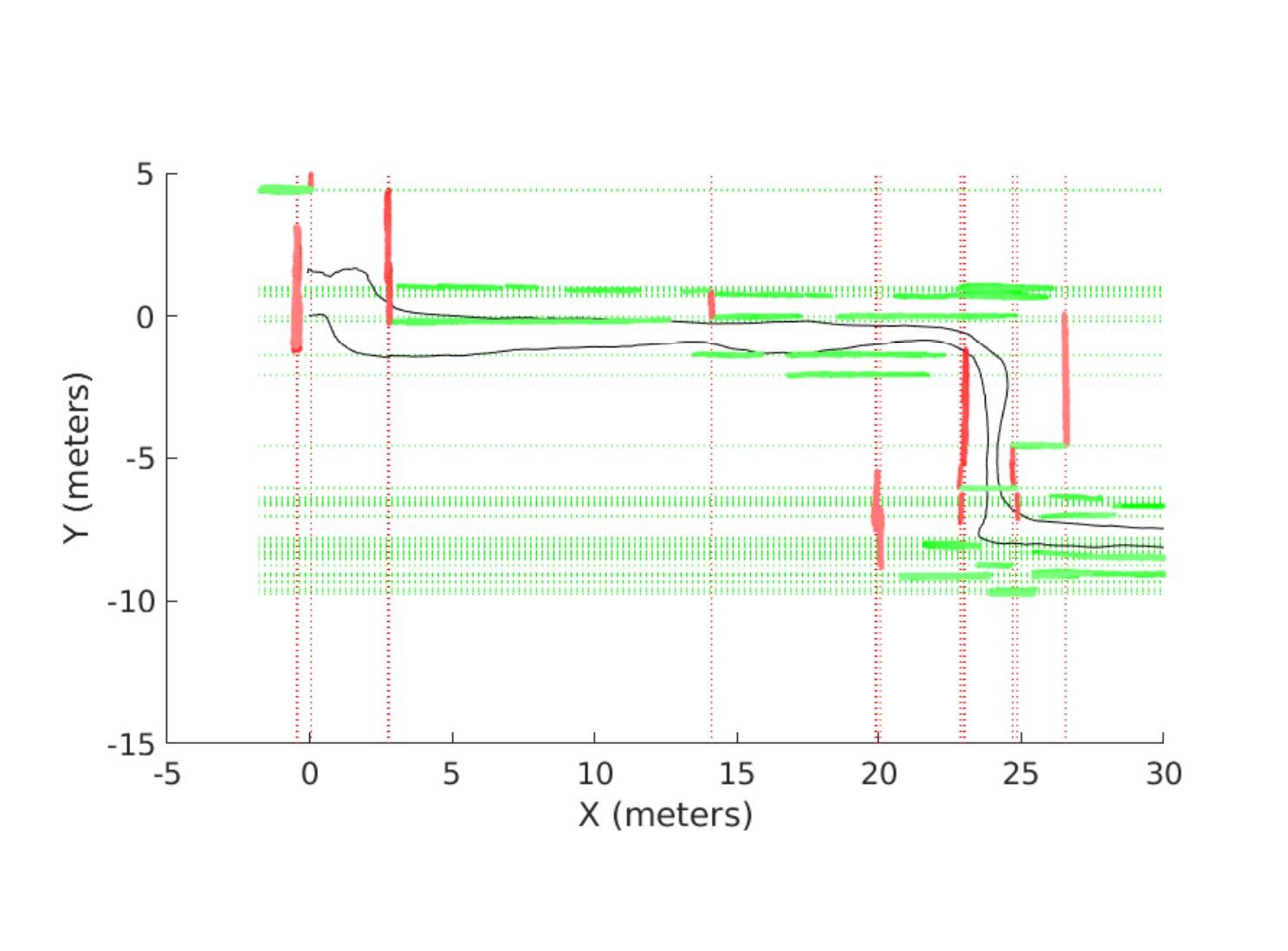}
    \includegraphics[width=0.7\columnwidth,clip,trim={0.5in 0.75in 0.75in 0.75in}]{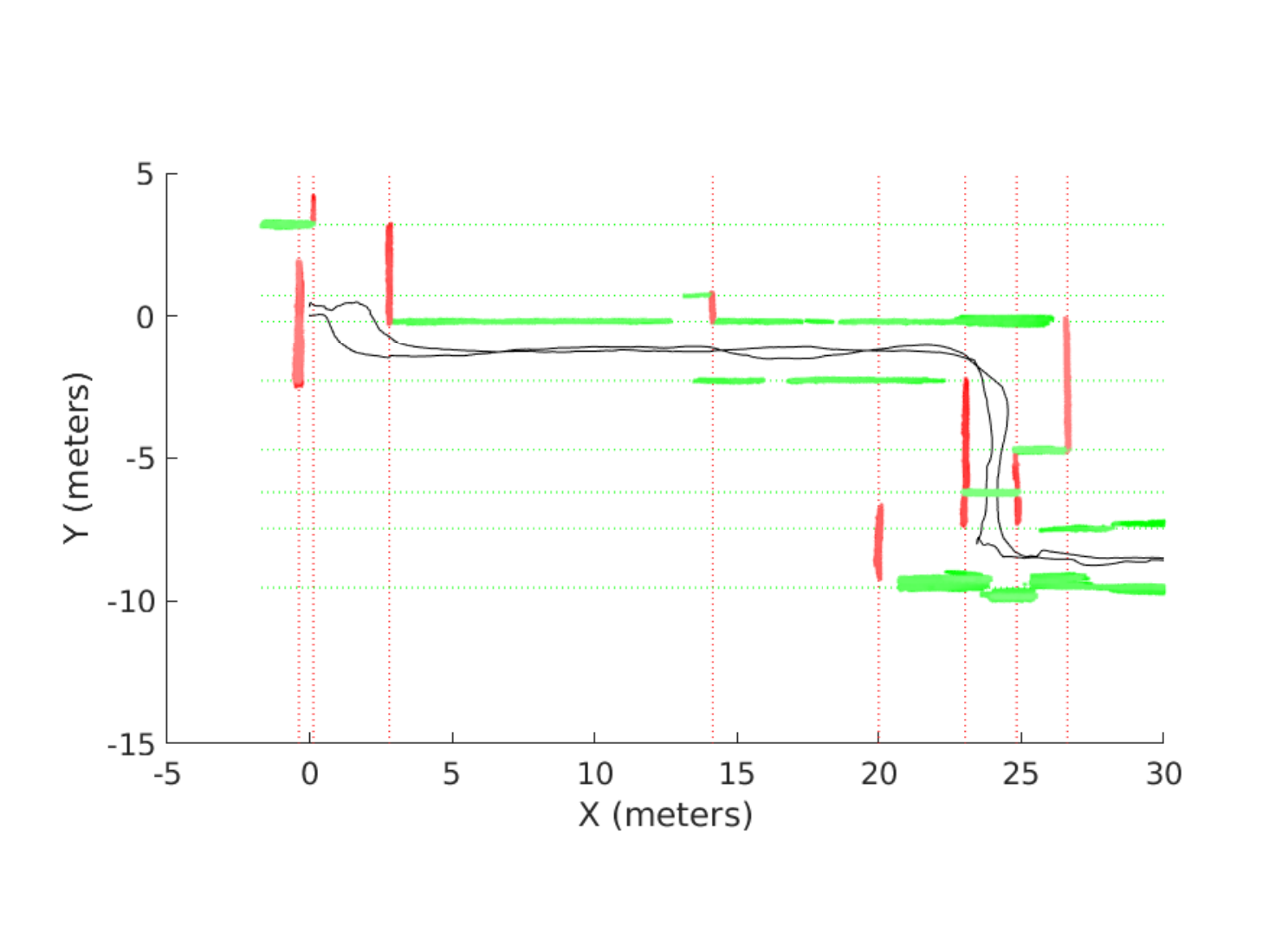}
    \caption{A birds-eye illustration of how our convex solution (below) can improve reconstruction by eliminating the drift still present in the least-squares solution (above). Notice the reduction in the total number of layout planes, which are denoted with red and green denoted lines corresponding to each axis-alignment. }
    \label{fig:comparison}
\end{figure}

Running this procedure yields the result shown in the second column of Figure \ref{fig:composite}. Each entry shows a result that captures the overall structure of the hallway but also exhibits the kind of drift typically associated with SLAM solutions; an artifact further highlighted in Figure \ref{fig:comparison}. These reconstruction errors stem from the fact that the initial set of correspondences derived from the stream of depth frames is necessarily incomplete. While correspondences derived from frame to frame analysis are typically correct they fail to capture salient long term matches. For example, when one enters then exits a room it is important to encode the fact that walls in the hallway were in fact previously seen and are not new features in the map. Similarly it is entirely possible to encounter a structural wall, then an opening, and then an entirely new section of the same wall. This problem of establishing long range correspondences is exacerbated by the fact that layout structures, unlike visual point feature landmarks, are extended structures and are rarely visually distinctive. Different sections of the same structure can have different appearances in different locations which can frustrate simple techniques that attempt to establish these long range correspondences.

We note that this problem of establishing long range correspondences subsumes the problem of loop closure which also revolves around the issue of deciding that one structure, a wall in this case, corresponds to one that was observed previously.

We propose a novel method that allows us to solve this problem by re-imagining this problem as one of model selection where our goal is to derive the simplest model that is consistent with our observations.

We begin by noting that solutions to Equation \ref{eq:linearopt} suffer from having too many wall surfaces. This is because when a layout structure is encountered again after an intervening break it will be entered again in the map as a new structural layout plane. Our goal then is to discover which of the segments in our overly large model could actually be coincident. Identifying two or more layout structures with each other effectively reduces the number of parameters associated with the model since all of the displacement parameters associated with that set are collapsed to a single value. In this way we effectively compress the model leading to a simpler solution.

We begin by encoding all of the possible or suspected equivalences between layout segments in a set of equations of the following form
\begin{equation}
m_a^x - m_b^x = 0
\label{eq:xequiv}
\end{equation}

As you would expect Equation \ref{eq:xequiv} encodes the idea that the $x$ aligned layout plane with index $a$ and the one with index $b$ are in fact the same. Analogous equations are defined for $y$ and $z$ aligned layout planes. 

These possible equivalences can be readily accumulated into a single sparse linear system $E \xi = 0$. Where $E$ is a sparse matrix encoding the relationship and $\xi$ is the vector of model parameters described earlier and used in Equation \ref{eq:linearopt}.

One possible approach to generating equivalence hypotheses, is to simply enumerate all possible equivalences between segments which face the same direction (north, east, south, west). However, this approach leads to an unnecessarily large $E$ matrix that contains numerous spurious hypotheses; the effects of which we discuss more thoroughly in Section \ref{sec:exp}. For now, we adopt the heuristic of enumerating all possible equivalences between segments facing in the same direction that are within some distance of each other. Depending on the length of the path, the expected amount of drift, and the initial number of planes detected, this value can vary between $0.5$-$3$ meters.

At this point we are not sure which of the equivalences are correct and which are false. This leads to a model selection problem. If there are $k$ possible equivalence relations then there are in principle $2^k$ possible models depending on which of the equivalence relations are enforced, modulo independence issues related to transitive closures among the equivalence relations.

How then can we go about selecting which relations are correct from this exponentially large set of possibilities?

We begin by using the original reconstruction problem as a system that defines a set of possible solutions. We do this by considering the set of $\xi$ values that satisfy:
\begin{equation}
\| A\xi - \mathbf{b} \|_2 \leq \delta
\end{equation}
where $\delta$ encodes the discrepancy between a proposed solution and the available measurements.

One way to choose delta is simply by setting it to 
\[(1 + \epsilon) \| A \xi^*_{\text{lin}} - \mathbf{b} \|_2, \]
where $\xi^*_{\text{lin}}$ is the optimal value of $\xi$ after solving Equation \ref{eq:linearopt}. Alternatively, one can relate $\delta$ to the error that one expects in the measurements based on the sensor model. We could also imagine replacing the $\ell_2$ norm with the $\ell_1$ or $\ell_\infty$ norms. In each case this inequality defines a convex set in parameter space corresponding to solutions that are sufficiently consistent with the original set of measurements.

We then view our problem as finding the point in the set that maximizes the number of equivalence relations we can satisfy. Note that maximizing the number of equivalences is equivalent to minimizing the number of parameters in the final model, so our goal is to effectively apply the principle of Occam’s razor to find the simplest model that explains our data.

Formally we can state our goal as follows
\begin{equation}
\begin{aligned}
& \underset{\xi}{\text{minimize}} && \| E \xi \|_0 \\
& \text{subject to} && \| A \xi - \mathbf{b} \|_2 \leq \delta\\
\end{aligned}
\end{equation}

In this expression, the $\ell_0$ norm of a vector simply counts the number of non-zero entries in its input. This problem formulation is reminiscent of the kinds of problems one encounters in compressed sensing.

While this formulation is what we would ideally like to tackle, the discontinuous nature of the $\ell_0$ norm makes it intractable so we resort instead to the $\ell_1$ norm which we can view as a convex relaxation of our original problem.

Our new goal then can be stated as follows
\begin{equation}
\begin{aligned}
& \underset{\xi}{\text{minimize}} && \| E \xi \|_1 \\
& \text{subject to} && \| A \xi - \mathbf{b} \|_2 \leq \delta\\
\end{aligned}
\label{eq:convexopt1}
\end{equation}

Many may notice the similarity between our formulation and the LASSO procedure \cite{Tibshirani2011}. LASSO performs subset selection over model coefficients by forcing as many of them to zero by bounding the sum of the absolute values of regression coefficients. Our approach to model simplification is different as our procedure reduces model complexity by enforcing equivalence relations encoded in the $E$ matrix.

At this point we note that the optimization problem stated in Equation \ref{eq:convexopt1} involves minimizing a convex function subject to a convex constraint which places us squarely in the domain of convex optimization. 
The resulting problem can be reformulated as solving for the optimal value of a linear objective function subject to a set of linear and convex quadratic constraints. We note that we can solve problems involving hundreds of variables in a matter of seconds due to the sparseness of the underlying systems. 
In our current implementation we formulate and solve this problem in Matlab using CVX.

Once the problem has been solved we examine the resulting vector $E \xi$ and apply a threshold $\mu$ to decide which of the equivalences should be enforced. We then re-solve the optimization problem enforcing these equivalences
\begin{equation}
\begin{aligned}
& \underset{\xi}{\text{minimize}} &&  \| A \xi - \mathbf{b} \|_2 \\
& \text{subject to} &&  E' \xi = 0\\
\end{aligned}
\label{eq:convexopt2}
\end{equation}
where $E'$ denotes the reduced set of enforced equivalences. The extent of the new layout structures are determined by computing the boundary around the individual corresponding layout segments residing on the same plane.

We also introduce an additional hard constraint on the trajectory of the sensor relative to the layout segments. For instance, a measurement $d_{ij}$ to layout segment $m_j^x$ also introduces the following linear inequality constraint
\begin{equation}
	-\mathbf{sign}(d_{ij}) (m_j^x - p_i^x) \leq 0
\end{equation}
This  constraint ensures that the recovered model is topologically consistent with the range observations. Accumulating these inequalities yields an additional convex constraint $D \xi \leq 0$, which is added to Equations \ref{eq:convexopt1} and \ref{eq:convexopt2}.


\section{Experimental Results}
\label{sec:exp}

\begin{table*}[t]
\caption{Complexity results in each of the mapped environments}
\label{tab:results}
\centering
\begin{tabular}{| l || c | c | c | c | c | c |}
\hline 
Area ID & 1 & 2 & 3 & 4 & 5 & 6 \\ \hline \hline
Number of Equivalences Considered & 151 & 2734 & 15631 & 454 & 1173 & 1704 \\ \hline
Number of Equivalences Accepted & 125 & 1431 & 10761 & 247 & 1092 & 1510 \\ \hline \hline
Number of Initial Layout Segments in Model & 48 & 239 & 408 & 89 & 108 & 167 \\ \hline
Number of Layout Structures After Analysis & 14 & 34 & 40 & 28 & 25 & 16 \\ \hline
Complexity Reduction \% & 70.8 & 85.8 & 90.2 & 68.5 & 76.9 & 90.4 \\ \hline \hline
Optimization Time (s) & 0.12 & 2.02 & 50.67 & 0.23 & 0.79 & 0.59 \\ \hline \hline
Path Length (m) & 53 & 200 & 249 & 69 & 113 & 67 \\ \hline
\end{tabular}
\end{table*}

\begin{table}
    \caption{Drift in meters after each type of optimization}
    \label{tab:driftcomp}
    \centering
    \begin{tabular}{|l|c|c|c|c|}
        \hline
        Optim. & Raw VIO & Entropy Compass & Least-Squares & Convex \\ \hline \hline
        Area 1 & 0.67 & 0.82 & 0.62 & 0.16 \\ \hline
        Area 2 & 2.36 & 3.06 & 1.58 & 0.39 \\ \hline
        Area 3 & 2.24 & 2.98 & 1.63 & 0.19 \\ \hline
        Area 4 & 0.46 & 0.35 & 0.48 & 0.09 \\ \hline
        Area 5 & 1.32 & 0.81 & 0.46 & 0.15 \\ \hline
        Area 6 & 0.73 & 1.07 & 0.88 & 1.11 \\ \hline
    \end{tabular}
\end{table}

Several reconstruction experiments were carried out in different areas of our campus. Three different sensor rigs were used in these experiments in order to demonstrate robustness. All of them have in common a custom stereo monochrome camera running at 20 Hz with a $97^{\circ} \times 81^{\circ}$ field of view (FOV) \cite{ovc}. Together with a hardware synchronized inertial measurement unit (IMU), they provide data to drive the stereo VIO algorithm. The primary difference between rigs is the choice of depth sensor and its frame rate. Rig A (used for Areas 1,2, and 3) hosts a PMD Monstar time-of-flight depth sensor, which captures $352 \times 287$ resolution frames at 10 Hz and has a FOV of $100^{\circ} \times 85^{\circ}$. Rig B (used for Areas 4 and 5) features the same depth sensor, but run at 5Hz. Lastly, Rig C (used for Area 6) replaces the Monstar with an Orbbec Astra structured light camera, which runs at 30Hz and has a resolution of $640 \times 400$ and a FOV of $60^{\circ} \times 49.5^{\circ}$.

With the exception of the threshold used for enumerating possible equivalences, all computations were carried out using the same set of optimization parameters agnostic to sensor choice or environment. The entropy compass explored a radius of $0.3^{\circ}$ around the estimate of angular displacement provided by the VIO system. The value of $\epsilon$ used to determine $\delta$ was empirically set to $2\%$. Finally, the threshold value of $\mu$ used to conclude equivalence between two segments was set to $30$ centimeters. All of our computations were carried out using an Intel i7-8700K CPU with 32GB of RAM. 

\begin{figure*}[p]
\begin{tabular}{ c | >{\centering\arraybackslash}m{1.7in} >{\centering\arraybackslash}m{1.7in} >{\centering\arraybackslash}m{1.7in} }
     Area ID & Axis-Aligned Frame Registration & Least-Squares Optimization & Convex Optimization \\ \hline
     \rule{0pt}{0.8in} Area 1 & \includegraphics[height=1.2in,clip,trim={3.5cm 0cm 4cm 0cm}]{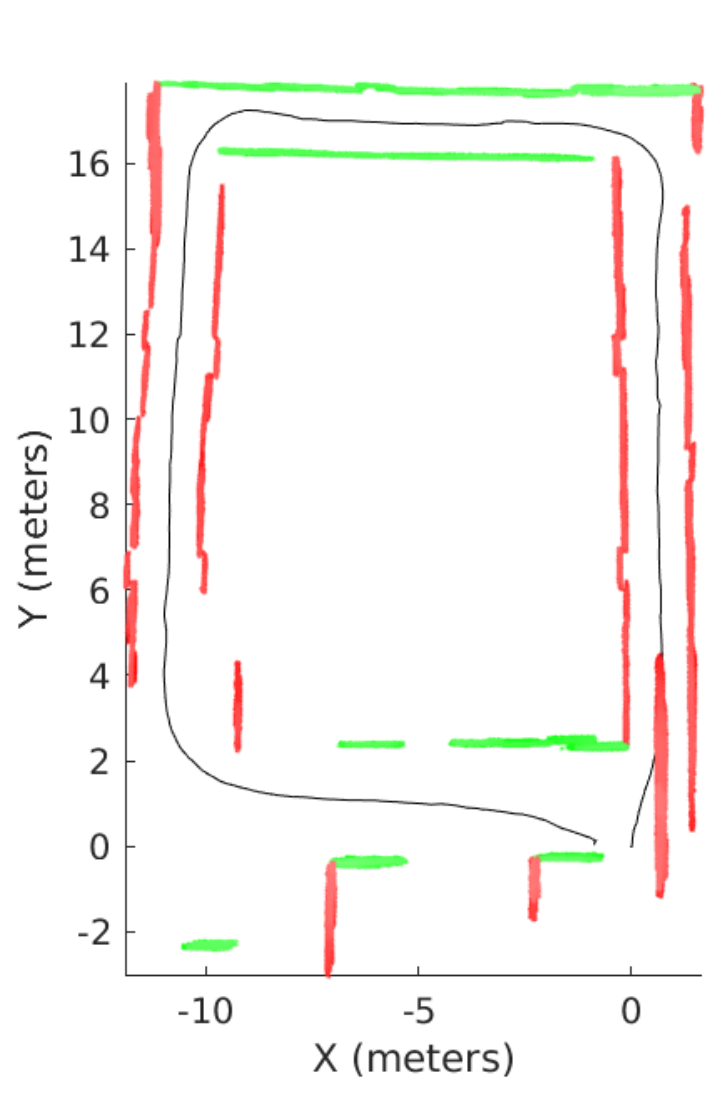} & \includegraphics[height=1.2in,clip,trim={3.5cm 0cm 4cm 0cm}]{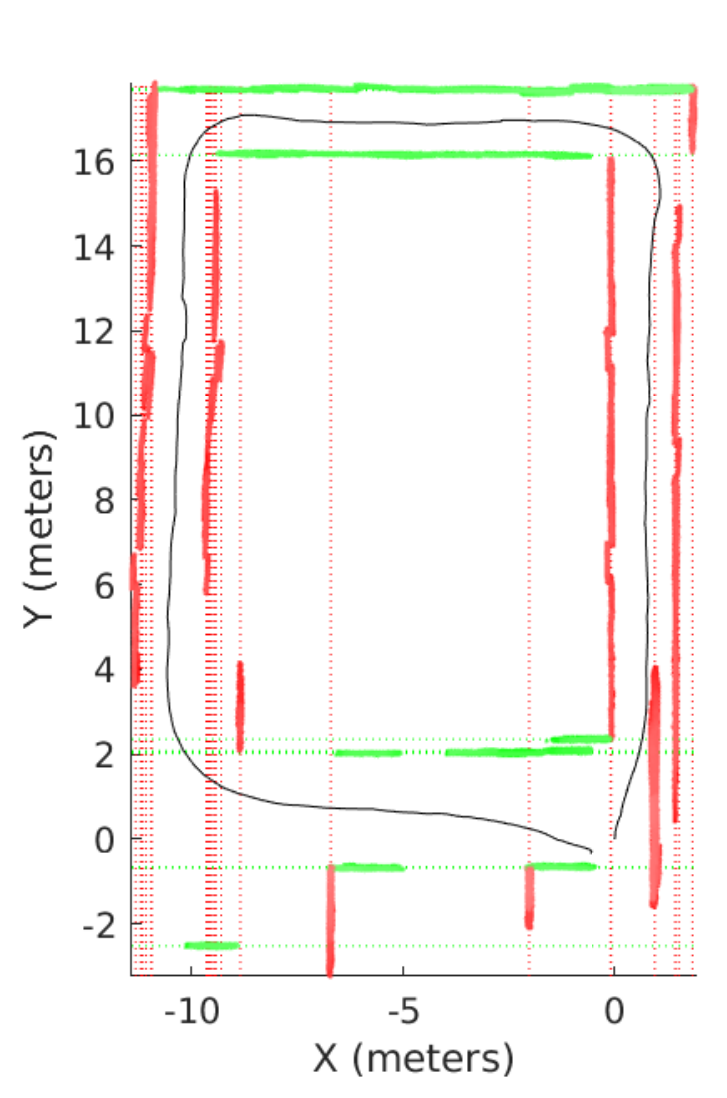} & \includegraphics[height=1.2in,clip,trim={3.5cm 0cm 4cm 0cm}]{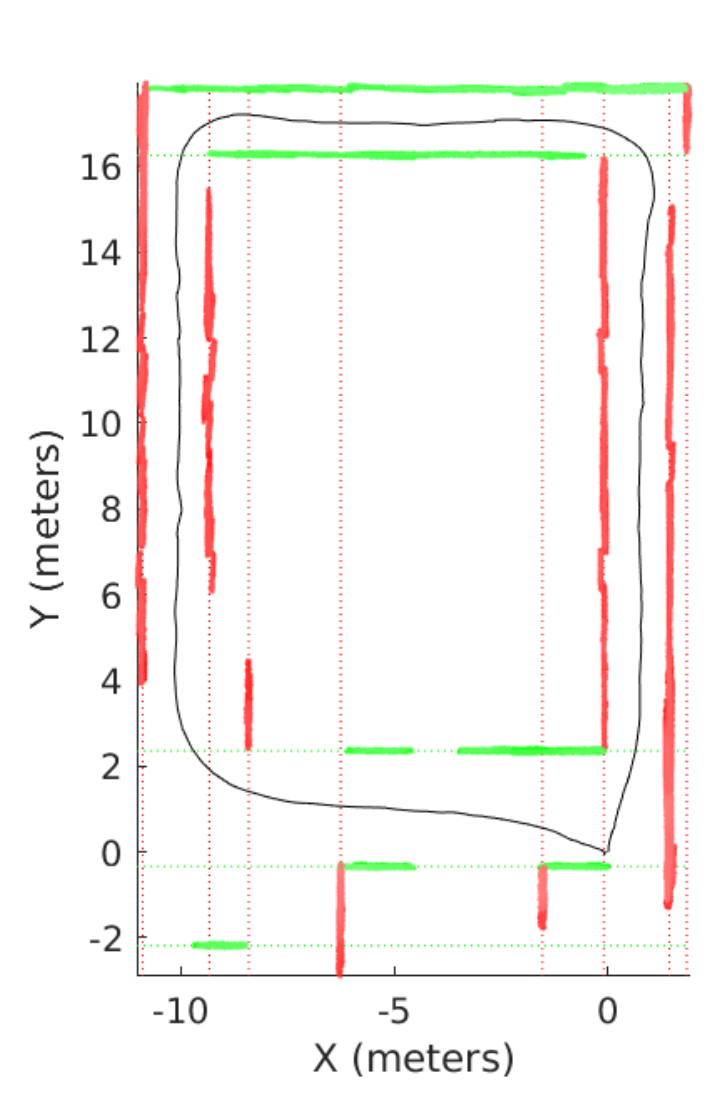} \\
     Area 2 & \includegraphics[height=0.95in,clip,trim={0.5cm 1.5cm 1cm 2cm}]{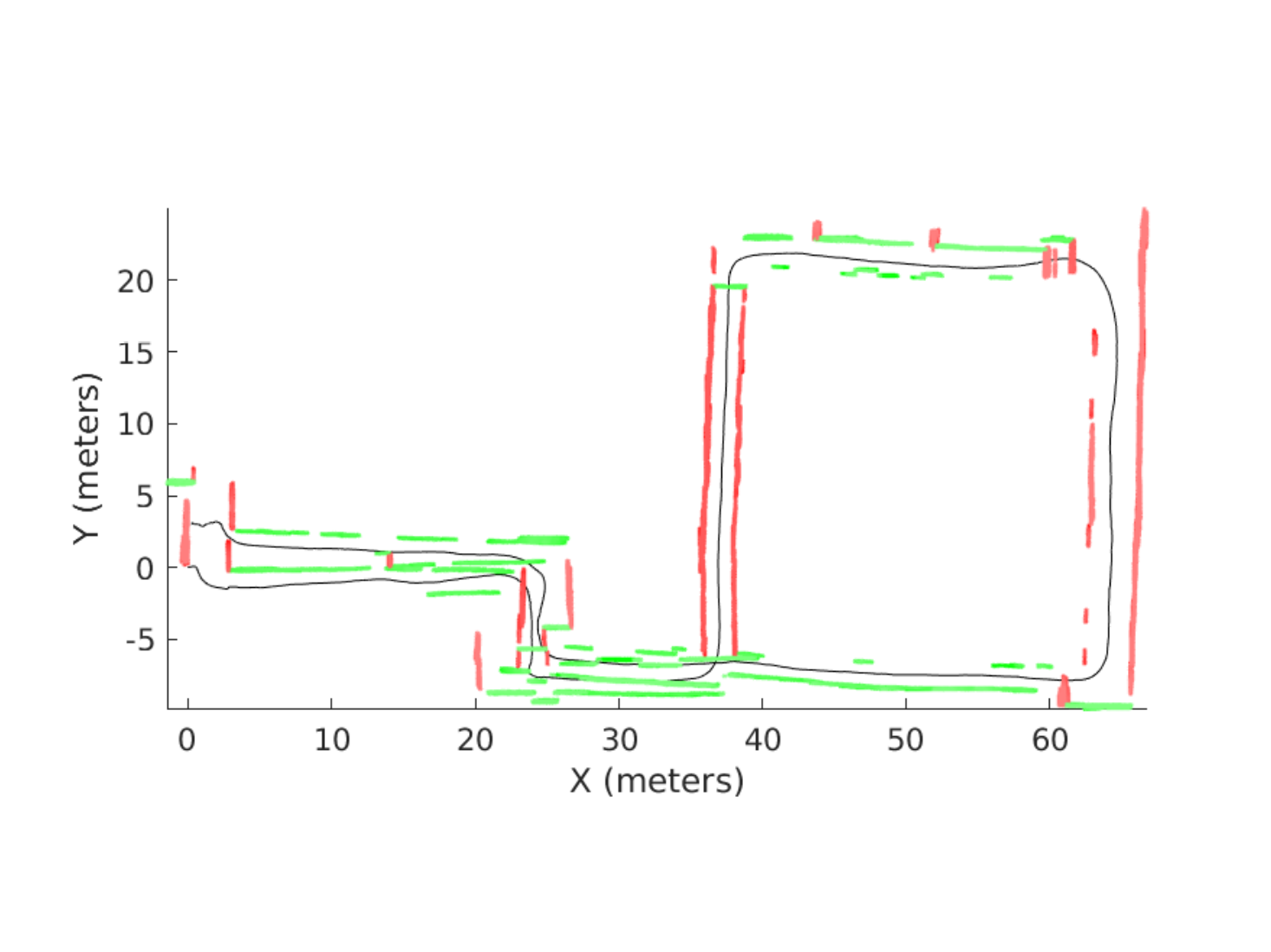} & \includegraphics[height=0.95in,clip,trim={0.5cm 1.5cm 1cm 2cm}]{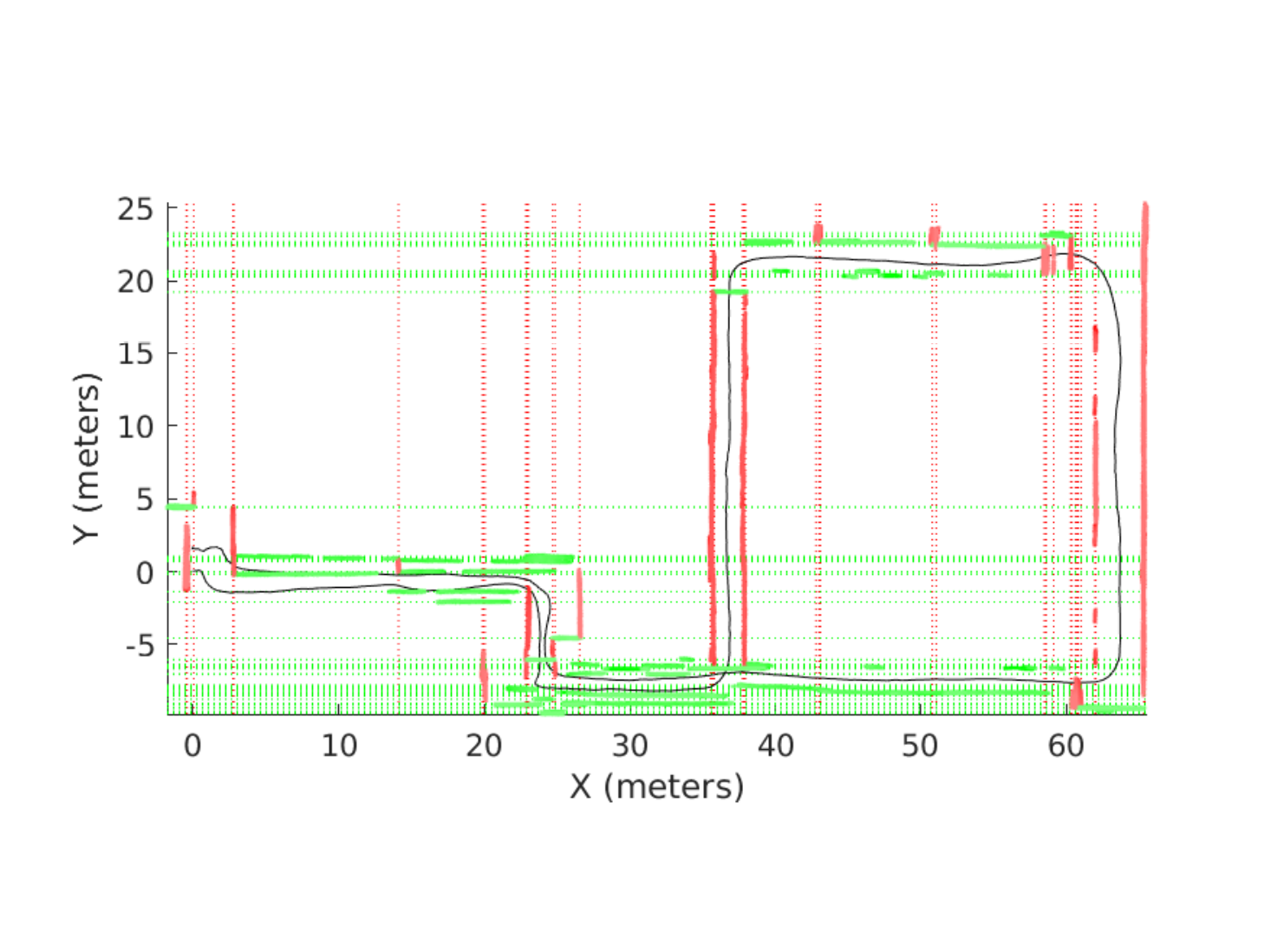} & \includegraphics[height=0.95in,clip,trim={0.5cm 1.5cm 1cm 2cm}]{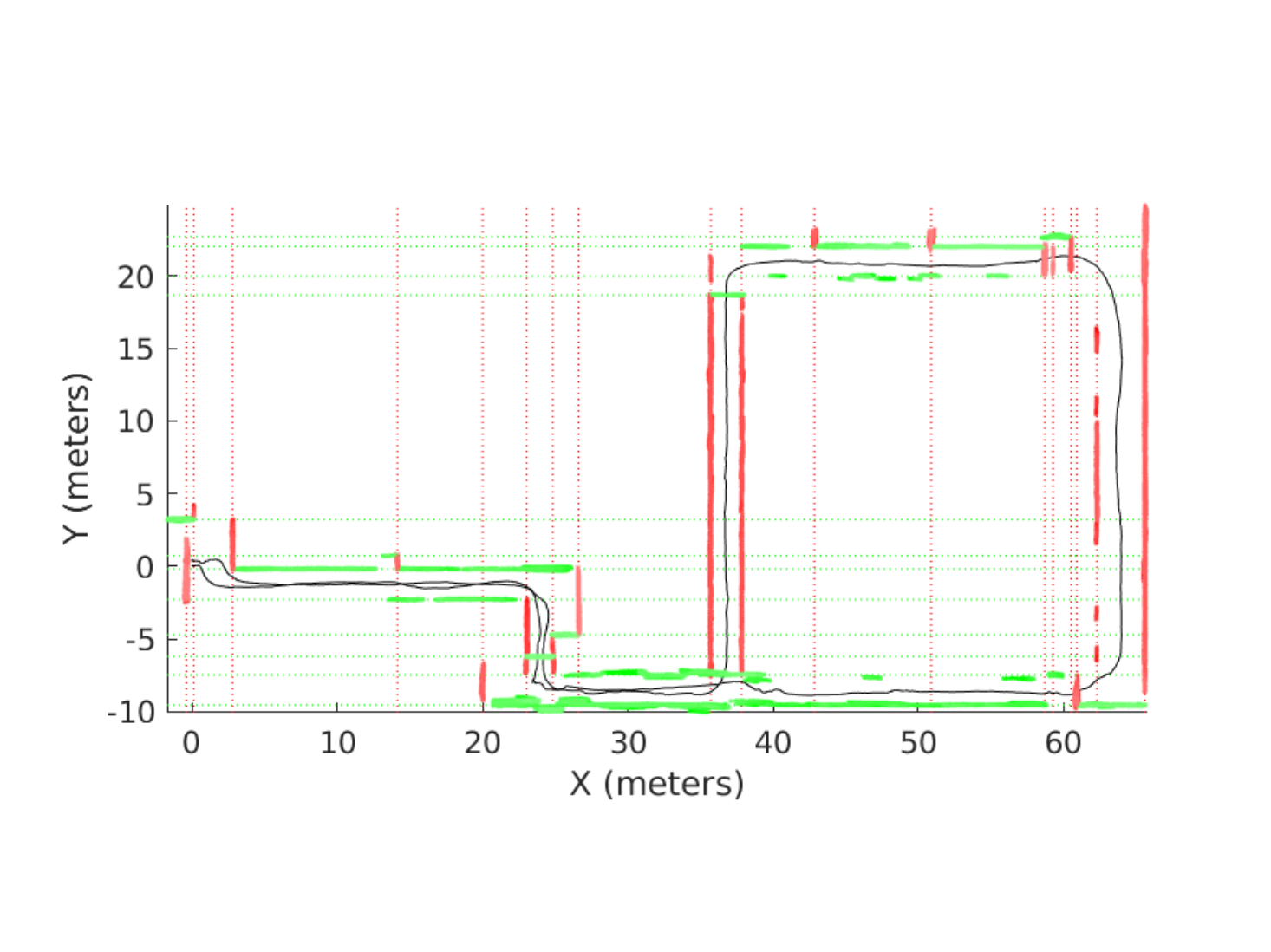} \\ 
     Area 3 & \includegraphics[height=2.2in,clip,trim={4cm 0cm 4.5cm 0.5cm}]{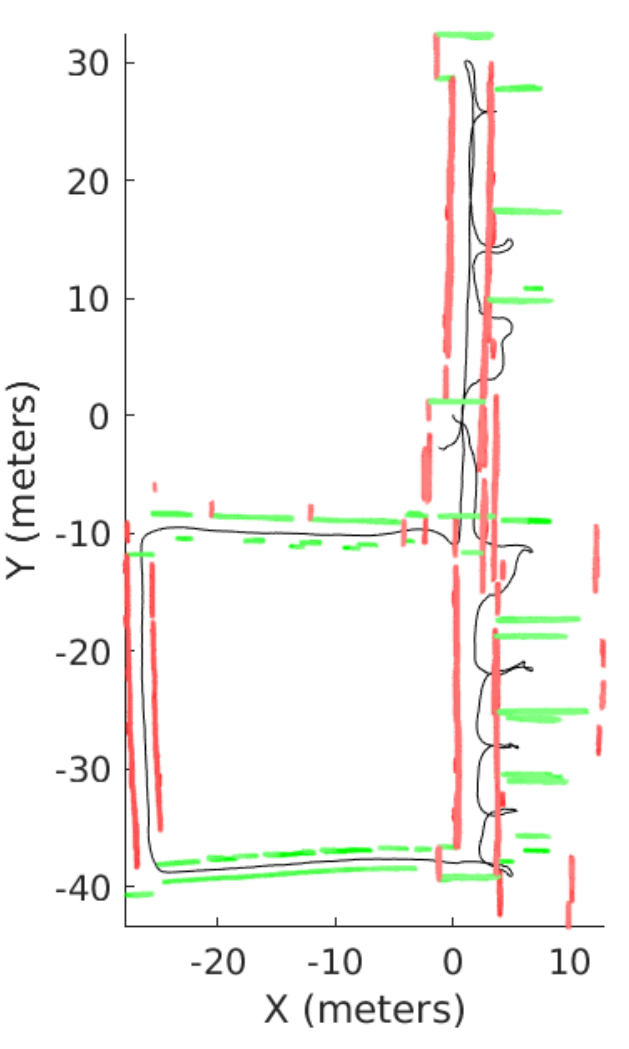} & \includegraphics[height=2.2in,clip,trim={4cm 0cm 4.5cm 0.5cm}]{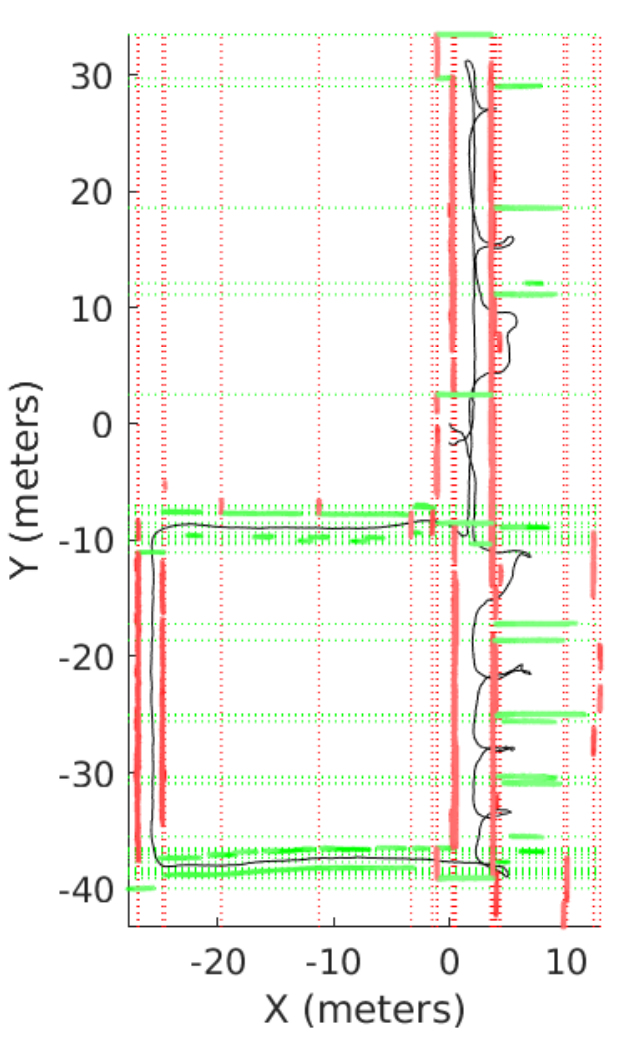} & \includegraphics[height=2.2in,clip,trim={4cm 0cm 4.5cm 0.5cm}]{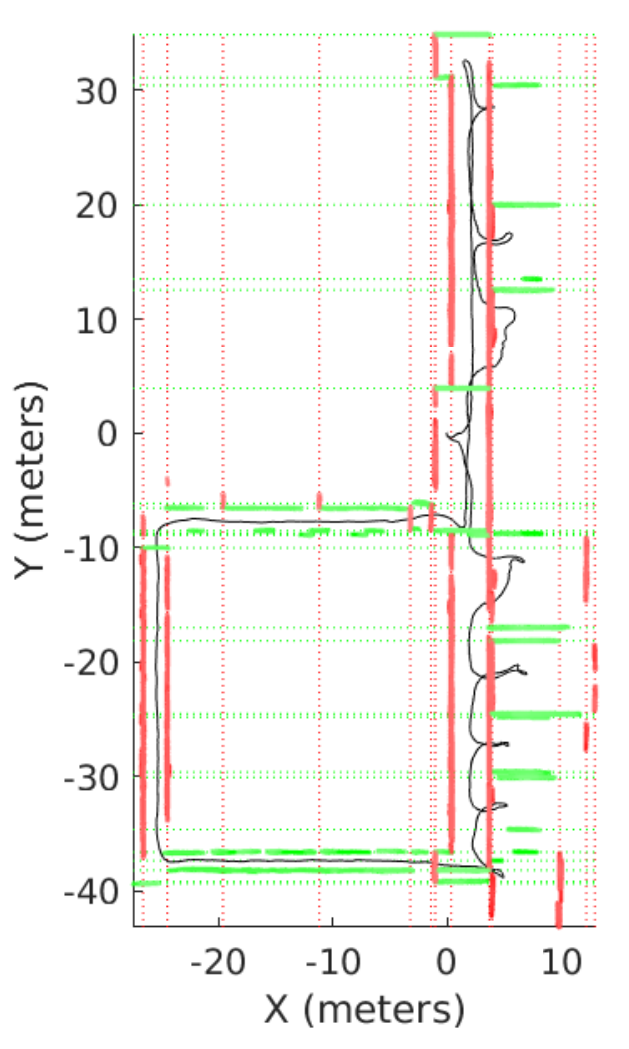} \\
     Area 4 & \includegraphics[height=1.5in,clip,trim={2cm 0cm 3cm 0.5cm}]{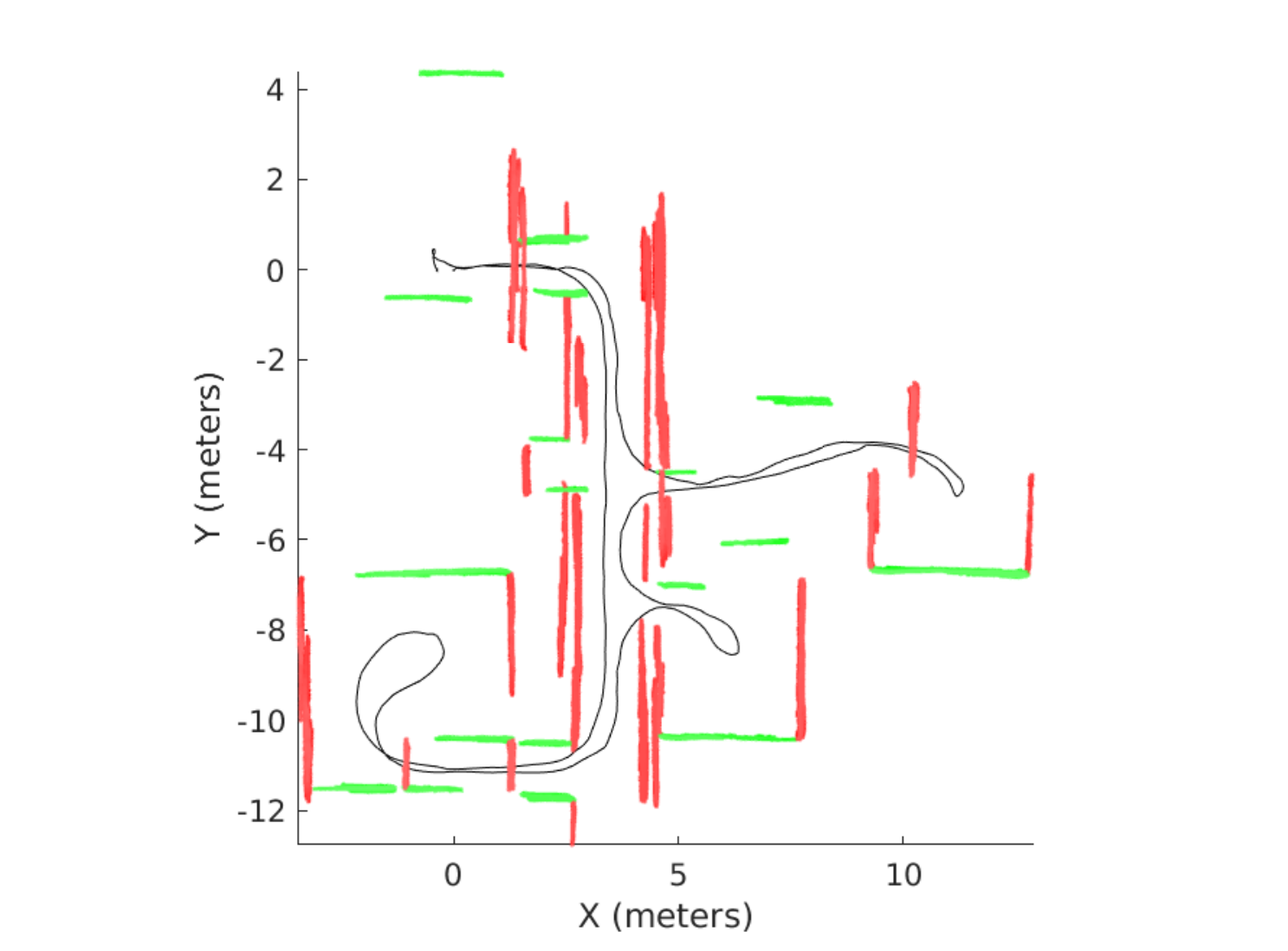} & \includegraphics[height=1.5in,clip,trim={2cm 0cm 3cm 0.5cm}]{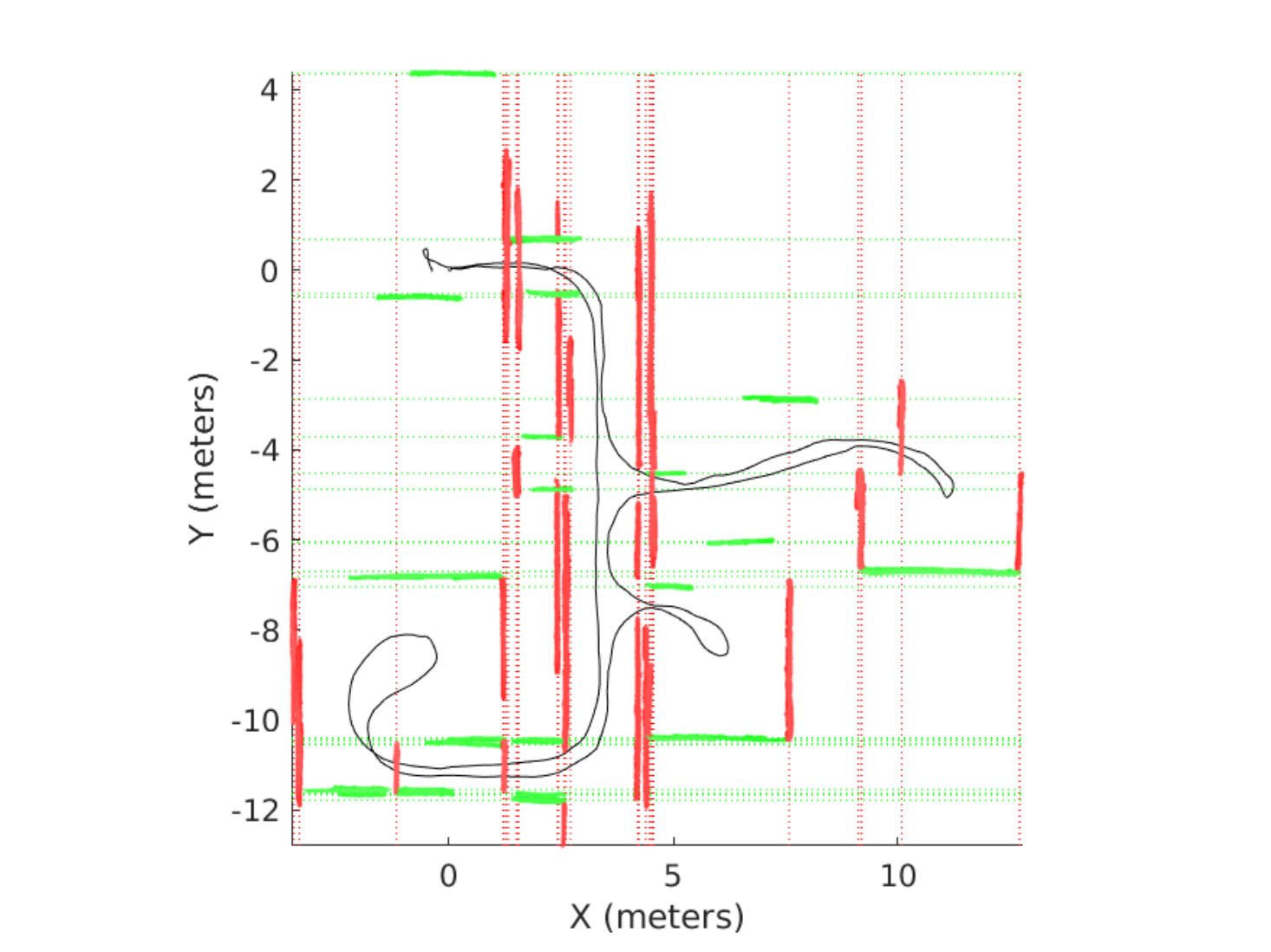} & \includegraphics[height=1.5in,clip,trim={2cm 0cm 3cm 0.5cm}]{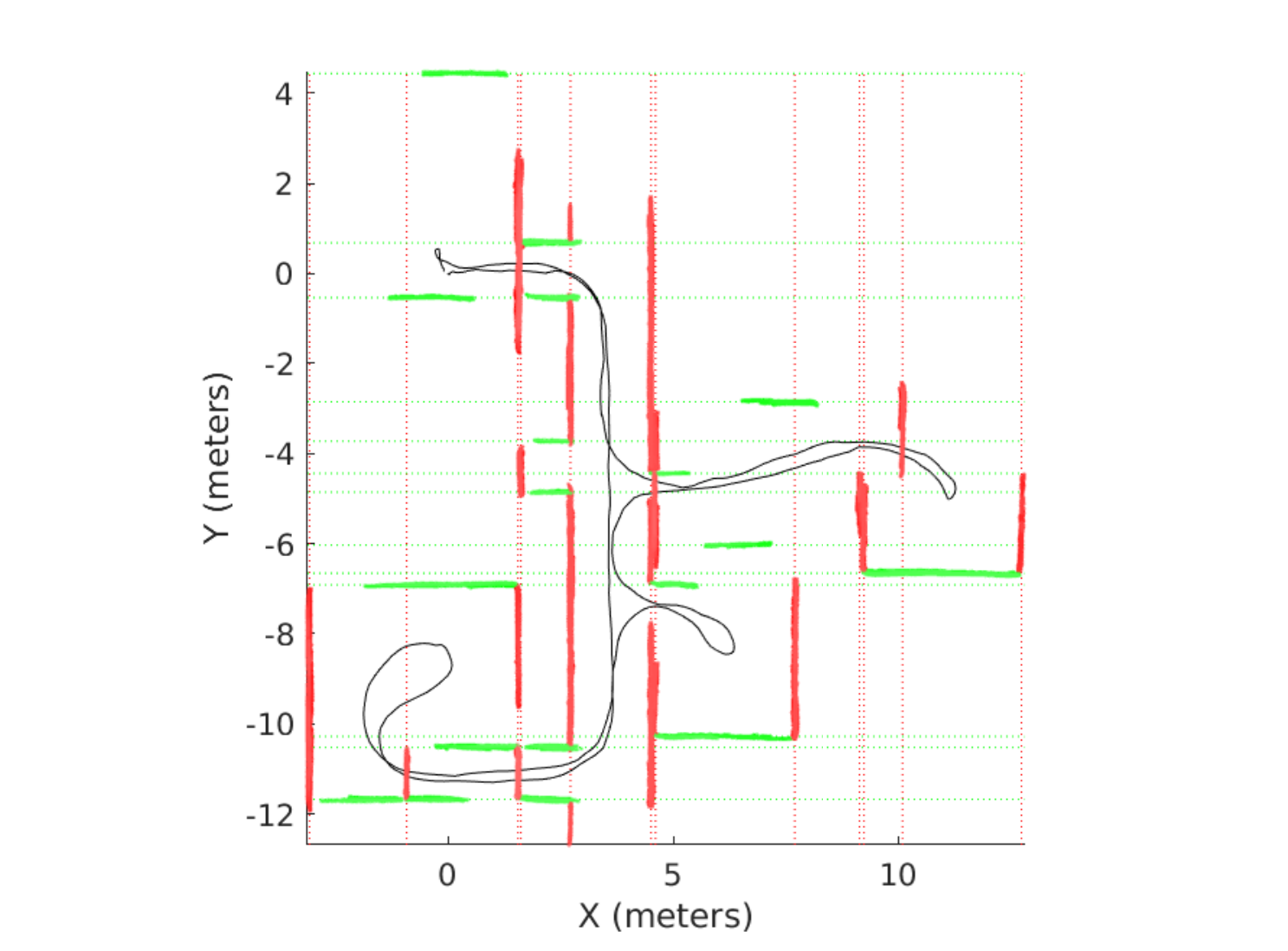} \\
     Area 5 & \includegraphics[height=1.4in,clip,trim={6cm 0cm 6.5cm 0.5cm}]{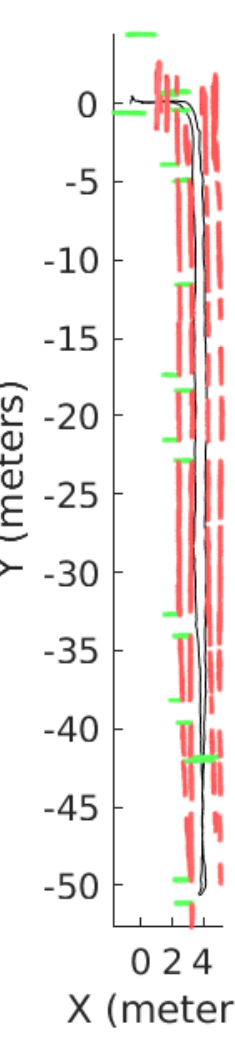} & \includegraphics[height=1.4in,clip,trim={6cm 0cm 6.5cm 0.5cm}]{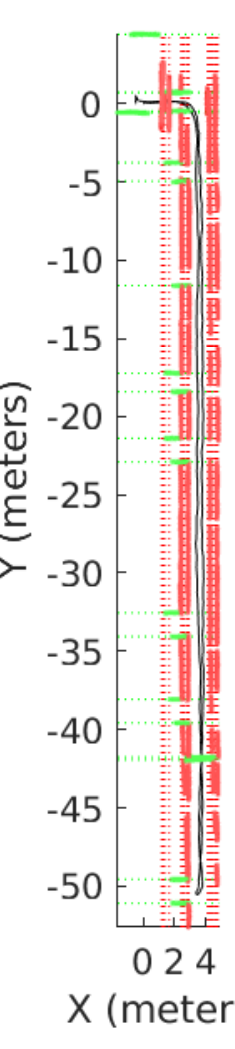} & \includegraphics[height=1.4in,clip,trim={6cm 0cm 6.5cm 0.5cm}]{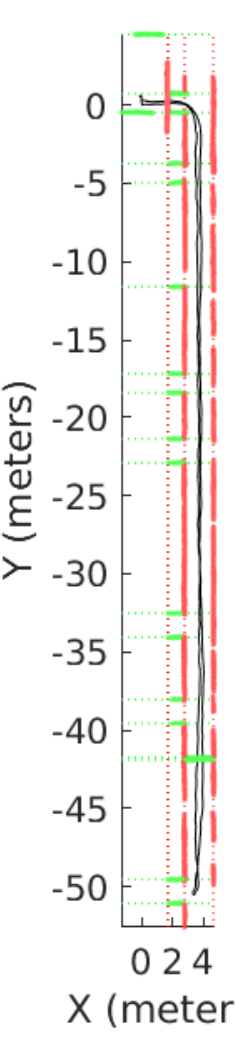} \\
     Area 6 & \includegraphics[height=0.8in,clip,trim={0.5cm 2cm 1cm 2.5cm}]{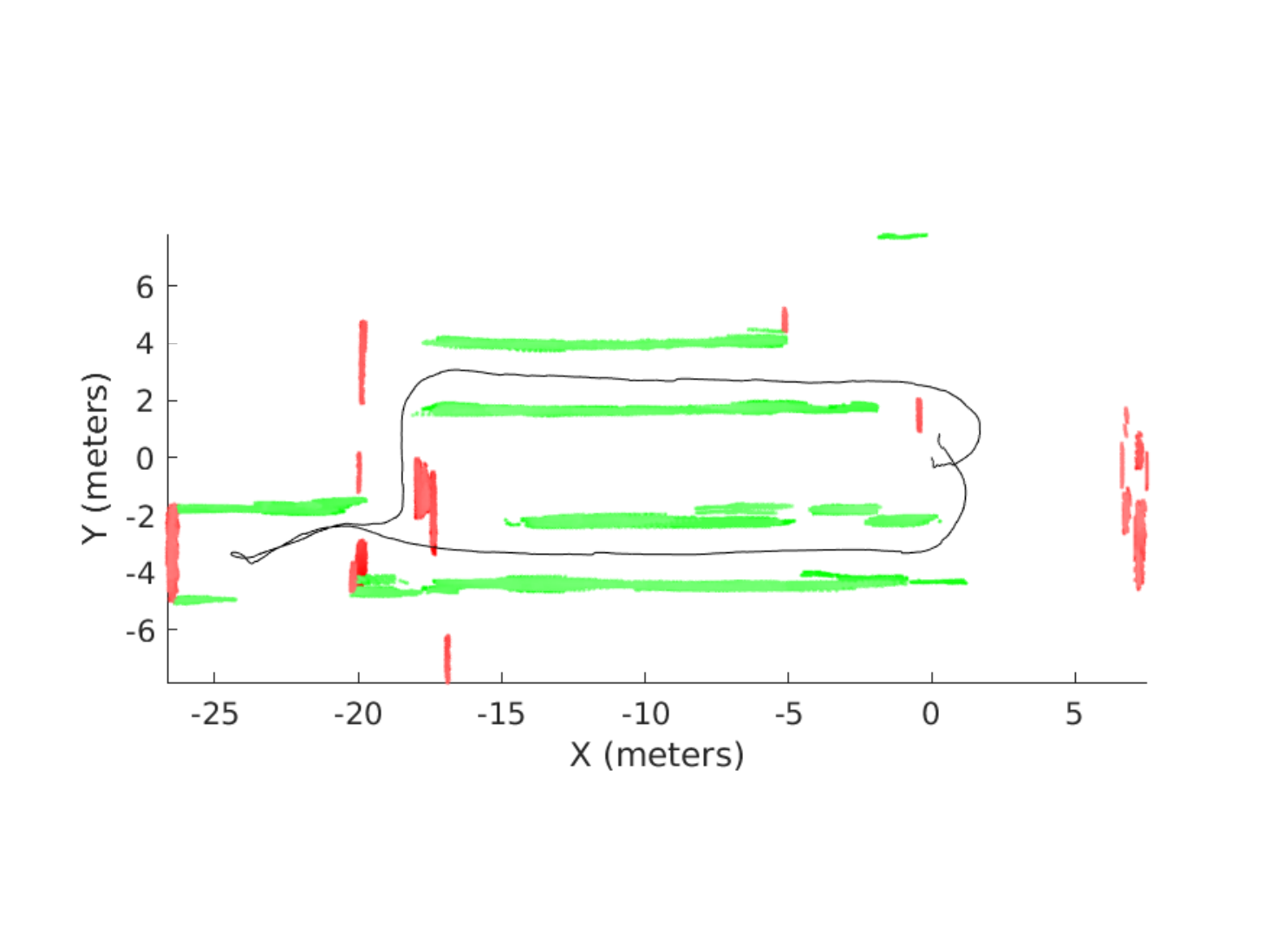} & \includegraphics[height=0.8in,clip,trim={0.5cm 2cm 1cm 2.5cm}]{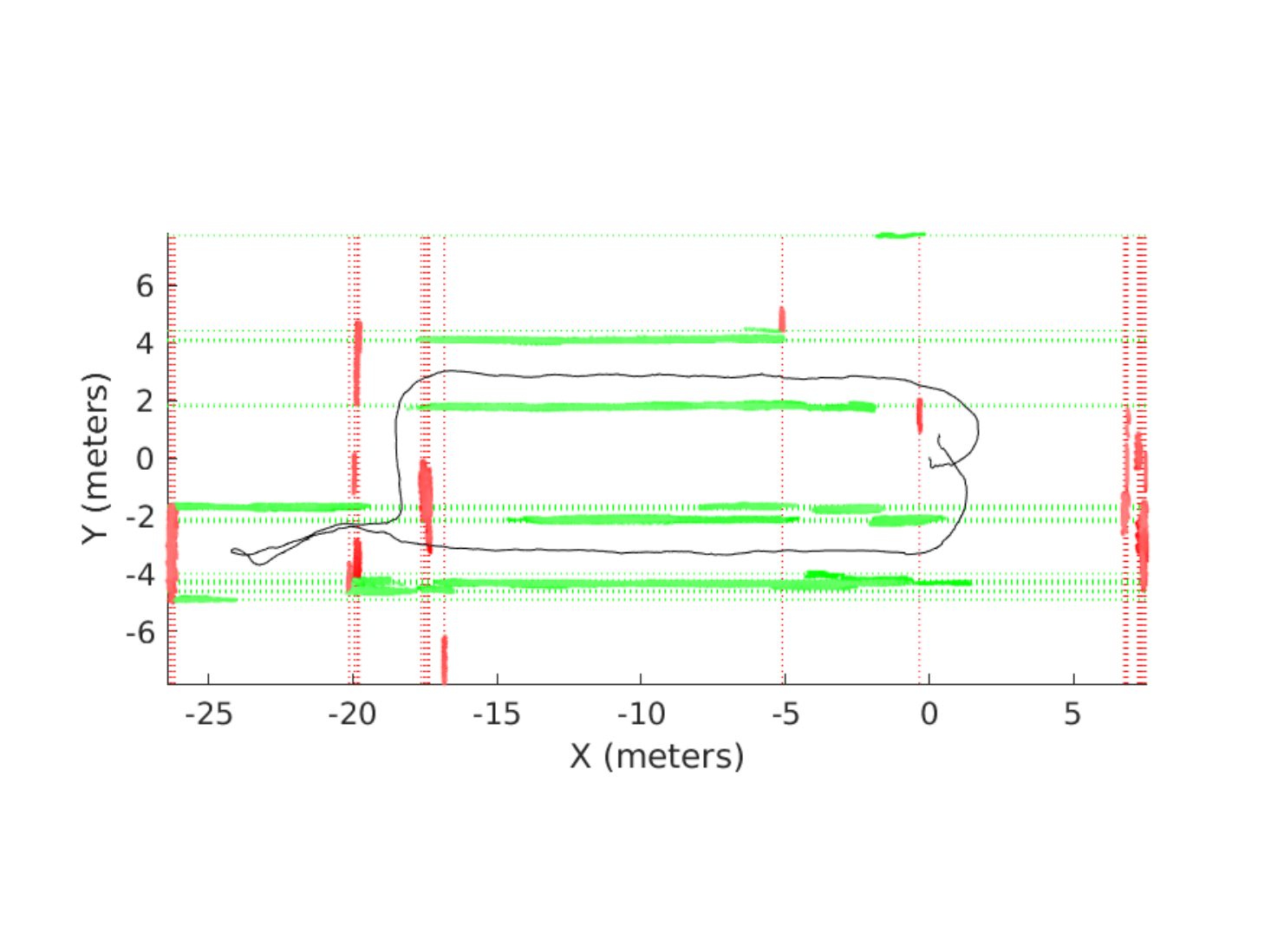} & \includegraphics[height=0.8in,clip,trim={0.5cm 2cm 1cm 2.5cm}]{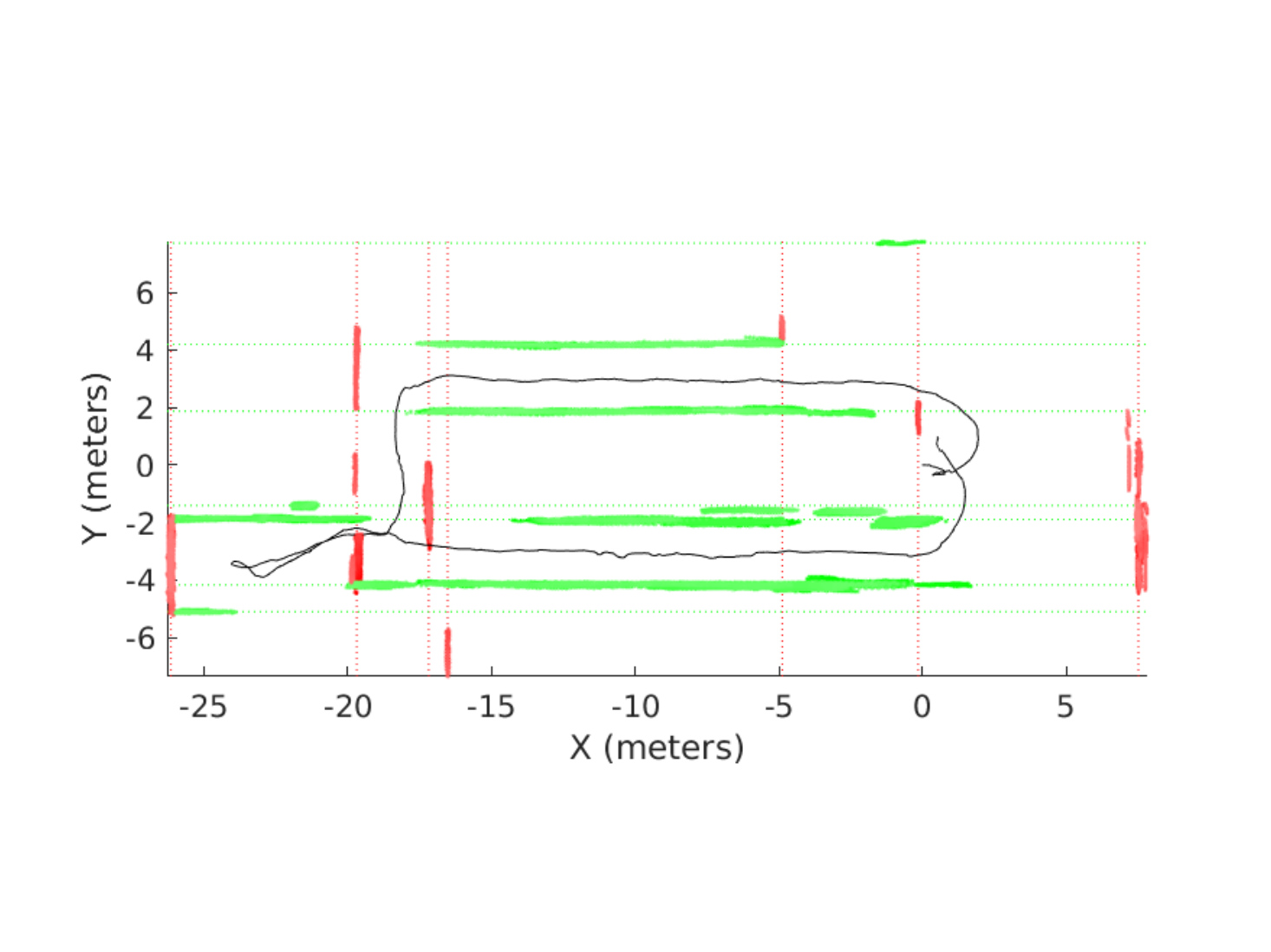} \\ 
\end{tabular}
\caption{Birds-eye view of the reconstruction results of our analysis in several Manhattan environments. Each column illustrates the effect of a different reconstruction method while each row corresponds to a different area. Red and green point clouds correspond to $x$ and $y$-aligned layout segments, which reside on infinite layout planes denoted in each figure with red and green dotted lines. The black curve illustrates the sensor trajectory. }
\label{fig:composite}
\end{figure*}

Figure \ref{fig:composite} illustrates a composite of the reconstruction results in each environment using our convex approach, which we compare to reconstructions based on registering layout segments to the axis-aligned frame position they were observed in, as well as the results of the least-squares optimization described in Equation \ref{eq:linearopt}.  

Different environments presented different challenges.
For instance, Area 3 is rather large and contains many classrooms along its eastern wing. As a result, redundant candidate layout structures were generated as we encountered the same walls multiple times upon entering and exiting each classroom, which our optimization then had to consider.
Area 6 features a modern architecture with many glass surfaces (embedded even in doors), large open areas, and exposed structural I-beams oriented at various angles. As a result, not only was the entropy analysis and layout segment detection confounded by the actual layout itself, but also by missing and corrupted depth measurements.
Importantly, almost all of the examples involve situations where the robot needs to perform loop closures to account for situations where the same surface is encountered again after a significant interval of time. These loop closures are automatically detected and factored into the reconstruction as part of our procedure.

For each environment, table \ref{tab:results} shows the number of equivalences considered, the number that were accepted, the number of layout structures in the original model and the number in the final simplified model. It also indicates the computational time required for the analysis and the length of the robot's trajectory. Note that the final optimized model contains far fewer layout planes than the original model.



Table \ref{tab:driftcomp} provides a quantitative analysis of the effects different types of optimization have on the trajectory drift. As the sensor rig is carried back to the starting location after each exploration, the values reported are the distances between the starting point and ending point of the trajectory after reconstruction.
Note that in all cases but one, the convex optimization significantly reduced the drift in the reconstructed trajectory.

We do note however that our system is not entirely immune to false data associations due to the use of a hard merge threshold. This can be problematic particularly in more complicated environments, such as Area 6, where multiple independent planes may occur within that threshold in a particular region and the error in the reconstructed trajectory may not be large enough to prevent an incorrect assignment. We are currently pursuing ideas such as weighted hypotheses and adaptive thresholds to address this issue.

\begin{figure}
    \centering
    \begin{subfigure}{0.45\columnwidth}
        \includegraphics[width=\columnwidth,clip,trim={4cm 0cm 4.5cm 0.5cm}]{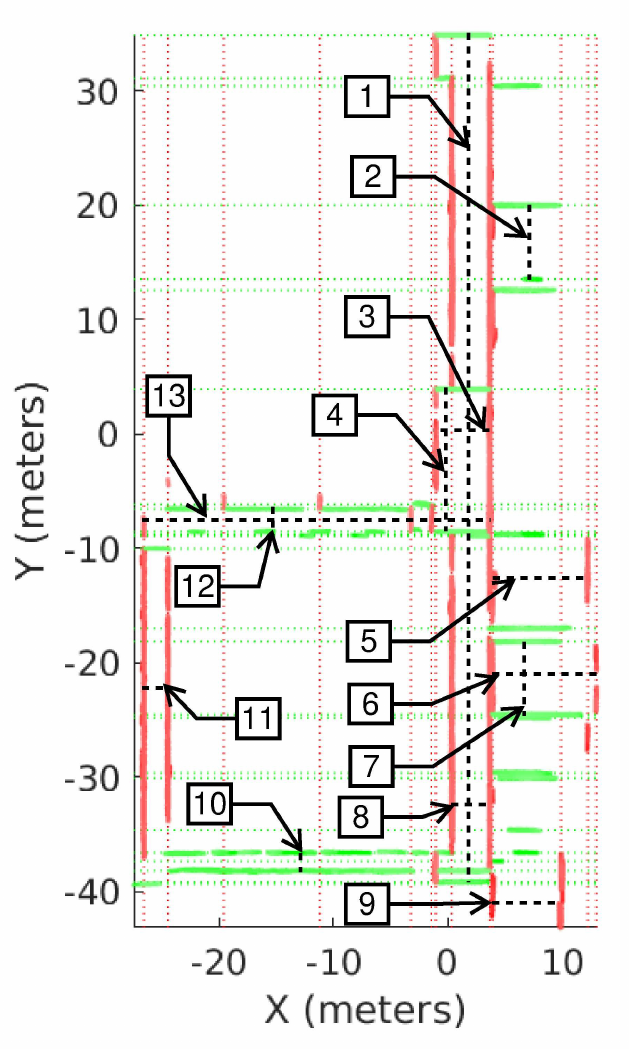}
    \end{subfigure}
    \begin{subfigure}{0.53\columnwidth}
        \begin{tabular}{|c|c|c|c|}
            \hline
             ID & GT & Model & $\Delta$  \\ \hline
             1 & 73.15 & 74.06 & 0.91 \\ \hline
             2 & 6.48 & 6.47 & 0.01 \\ \hline
             3 & 4.72 & 4.72 & 0 \\ \hline
             4 & 12.70 & 12.46 & 0.24 \\ \hline
             5 & 8.48 & 8.34 & 0.14 \\ \hline
             6 & 9.01 & 9.08 & 0.07 \\ \hline
             7 & 6.34 & 6.38 & 0.04 \\ \hline
             8 & 3.35 & 3.32 & 0.03 \\ \hline
             9 & 6.10 & 6.00 & 0.10 \\ \hline
             10 & 1.47 & 1.59 & 0.12 \\ \hline
             11 & 2.10 & 2.10 & 0 \\ \hline
             12 & 2.32 & 2.36 & 0.04 \\ \hline
             13 & 30.00 & 30.26 & 0.26 \\ \hline
        \end{tabular}
    \end{subfigure}
    \caption{Comparison of surface to surface distances against ground truth measurements collected with a laser range finder. }
    \label{fig:errcomp}
\end{figure}

Figure \ref{fig:errcomp} provides a quantitative analysis of the distance between selected surfaces in the recovered model compared to ground truth measurements of these distances taken with a laser range finder in Area 3. As these results demonstrate the average reconstruction error in this set of measurements is $1.5\%$.

\section{Conclusion}

In conclusion, we have demonstrated an approach for generating compact reconstructions of Manhattan environments. In scenarios where a reasonable estimate for one's rotation can be inferred from the visible Manhattan structure, we can solve the full SLAM problem using convex optimization. Furthermore, our sparse objective enables us to explore the vast combinatorial space of potential data associations and loop closures, which results in a more accurate trajectory alongside a compact representation of the map.
We validate our mapping procedure on a set of representative indoor environments.

\bibliographystyle{IEEEtran}
\bibliography{Loop_Closure,Scene_Parsing,SLAM,Manhattan_Frame_Estimation,Floor_Plan_Generation,misc}

\begin{thebibliography}{10}
\providecommand{\url}[1]{#1}
\csname url@samestyle\endcsname
\providecommand{\newblock}{\relax}
\providecommand{\bibinfo}[2]{#2}
\providecommand{\BIBentrySTDinterwordspacing}{\spaceskip=0pt\relax}
\providecommand{\BIBentryALTinterwordstretchfactor}{4}
\providecommand{\BIBentryALTinterwordspacing}{\spaceskip=\fontdimen2\font plus
\BIBentryALTinterwordstretchfactor\fontdimen3\font minus
  \fontdimen4\font\relax}
\providecommand{\BIBforeignlanguage}[2]{{%
\expandafter\ifx\csname l@#1\endcsname\relax
\typeout{** WARNING: IEEEtran.bst: No hyphenation pattern has been}%
\typeout{** loaded for the language `#1'. Using the pattern for}%
\typeout{** the default language instead.}%
\else
\language=\csname l@#1\endcsname
\fi
#2}}
\providecommand{\BIBdecl}{\relax}
\BIBdecl

\bibitem{Hsiao2018}
M.~Hsiao, E.~Westman, and M.~Kaess, ``{Dense Planar-Inertial SLAM with
  Structural Constraints},'' in \emph{IEEE International Conference on Robotics
  and Automation (ICRA)}, 2018, pp. 6521--6528.

\bibitem{Pumarola2017}
A.~Pumarola, A.~Vakhitov, A.~Agudo, A.~Sanfeliu, and F.~Moreno-Noguer,
  ``{PL-SLAM: Real-time monocular visual SLAM with points and lines},'' in
  \emph{IEEE International Conference on Robotics and Automation (ICRA)}, 2017.

\bibitem{Li2018}
H.~Li, J.~Yao, J.-c. Bazin, X.~Lu, Y.~Xing, and K.~Liu, ``{A Monocular SLAM
  System Leveraging Structural Regularity in Manhattan World},'' in \emph{IEEE
  International Conference on Robotics and Automation (ICRA)}, 2018, pp.
  2518--2525.

\bibitem{Li2017}
H.~Li, J.~Yao, X.~Lu, and J.~Wu, ``{Combining Points and Lines for Camera Pose
  Estimation and Optimization in Monocular Visual Odometry *},'' in
  \emph{IEEE/RSJ International Conference on Intelligent Robots and Systems
  (IROS)}, 2017, pp. 1289--1296.

\bibitem{Mur-Artal2015a}
R.~Mur-Artal, J.~M. Montiel, and J.~D. Tardos, ``{ORB-SLAM: A Versatile and
  Accurate Monocular SLAM System},'' \emph{IEEE Transactions on Robotics},
  vol.~31, no.~5, pp. 1147--1163, 2015.

\bibitem{Camposeco2015}
F.~Camposeco and M.~Pollefeys, ``{Using vanishing points to improve
  visual-inertial odometry},'' in \emph{IEEE International Conference on
  Robotics and Automation (ICRA)}, vol. 2015-June, no. June, 2015, pp.
  5219--5225.

\bibitem{Straub2017}
J.~{Straub}, R.~{Cabezas}, J.~{Leonard}, and I.~{Fisher}, John~W.,
  ``{Direction-Aware Semi-Dense SLAM},'' \emph{arXiv e-prints}, p.
  arXiv:1709.05774, Sep. 2017.

\bibitem{Zhou2015}
H.~Zhou, D.~Zou, L.~Pei, R.~Ying, P.~Liu, and W.~Yu, ``{StructSLAM: Visual SLAM
  with building structure lines},'' \emph{IEEE Transactions on Vehicular
  Technology}, vol.~64, no.~4, pp. 1364--1375, 2015.

\bibitem{Kim2018}
P.~Kim, B.~Coltin, and H.~J. Kim, ``{Low-Drift Visual Odometry in Structured
  Environments by Decoupling Rotational and Translational Motion},'' in
  \emph{IEEE International Conference on Robotics and Automation (ICRA)}, 2018,
  pp. 7247--7253.

\bibitem{Kim}
------, ``{Linear RGB-D SLAM for Planar Environments},'' in \emph{European
  Conference on Computer Vision (ECCV)}, 2018, pp. 1--16.

\bibitem{Bowman2017a}
S.~L. Bowman, N.~Atanasov, K.~Daniilidis, and G.~J. Pappas, ``{Probabilistic
  data association for semantic SLAM},'' in \emph{IEEE International Conference
  on Robotics and Automation (ICRA)}, 2017, pp. 1722--1729.

\bibitem{Lowry2016}
S.~Lowry, N.~Sunderhauf, P.~Newman, J.~J. Leonard, D.~Cox, P.~Corke, and M.~J.
  Milford, ``{Visual Place Recognition: A Survey},'' \emph{IEEE Transactions on
  Robotics}, vol.~32, no.~1, pp. 1--19, 2016.

\bibitem{Latif2017}
Y.~Latif, G.~Huang, J.~Leonard, and J.~Neira, ``{Sparse optimization for robust
  and efficient loop closing},'' \emph{Robotics and Autonomous Systems},
  vol.~93, pp. 13--26, 2017.

\bibitem{Okorn2010}
B.~Okorn, X.~Xiong, B.~Akinci, and D.~Huber, ``{Toward Automated Modeling of
  Floor Plans},'' in \emph{Symposium on 3D Data Processing, Visualization and
  Transmission}, 2010.

\bibitem{Mura2016}
\BIBentryALTinterwordspacing
C.~Mura, O.~Mattausch, and R.~Pajarola, ``{Piecewise-planar Reconstruction of
  Multi-room Interiors with Arbitrary Wall Arrangements},'' \emph{Computer
  Graphics Forum}, vol.~35, no.~7, pp. 179--188, oct 2016. [Online]. Available:
  \url{http://doi.wiley.com/10.1111/cgf.13015}
\BIBentrySTDinterwordspacing

\bibitem{Ochmann2016}
S.~Ochmann, R.~Vock, R.~Wessel, and R.~Klein, ``{Automatic reconstruction of
  parametric building models from indoor point clouds},'' \emph{Computers and
  Graphics (Pergamon)}, vol.~54, pp. 94--103, feb 2016.

\bibitem{Liu2018a}
\BIBentryALTinterwordspacing
C.~Liu, J.~Wu, and Y.~Furukawa, ``Floornet: {A} unified framework for floorplan
  reconstruction from 3d scans,'' \emph{CoRR}, vol. abs/1804.00090, 2018.
  [Online]. Available: \url{http://arxiv.org/abs/1804.00090}
\BIBentrySTDinterwordspacing

\bibitem{Angladon2018}
V.~Angladon, S.~Gasparini, and V.~Charvillat, ``{Room floor plan generation on
  a project tango device},'' in \emph{Lecture Notes in Computer Science
  (including subseries Lecture Notes in Artificial Intelligence and Lecture
  Notes in Bioinformatics)}, vol. 10705 LNCS, 2018, pp. 226--238.

\bibitem{Cabral2014}
R.~Cabral and Y.~Furukawa, ``{Piecewise planar and compact floorplan
  reconstruction from images},'' in \emph{IEEE Conference on Computer Vision
  and Pattern Recognition (CVPR)}, 2014.

\bibitem{Sun2017a}
\BIBentryALTinterwordspacing
K.~Sun, K.~Mohta, B.~Pfrommer, M.~Watterson, S.~Liu, Y.~Mulgaonkar, C.~J.
  Taylor, and V.~Kumar, ``{Robust Stereo Visual Inertial Odometry for Fast
  Autonomous Flight},'' \emph{IEEE Robotics and Automation Letters (RA-L)},
  2017. [Online]. Available: \url{http://arxiv.org/abs/1712.00036}
\BIBentrySTDinterwordspacing

\bibitem{Bazin2013}
\BIBentryALTinterwordspacing
J.~C. Bazin, Y.~Seo, and M.~Pollefeys, ``{Globally optimal consensus set
  maximization through rotation search},'' in \emph{Lecture Notes in Computer
  Science (including subseries Lecture Notes in Artificial Intelligence and
  Lecture Notes in Bioinformatics)}, vol. 7725 LNCS, no. PART 2.\hskip 1em plus
  0.5em minus 0.4em\relax Springer, Berlin, Heidelberg, 2013, pp. 539--551.
  [Online]. Available:
  \url{http://link.springer.com/10.1007/978-3-642-37444-9{\_}42}
\BIBentrySTDinterwordspacing

\bibitem{Taneja2015}
\BIBentryALTinterwordspacing
A.~Taneja, L.~Ballan, and M.~Pollefeys, ``{Never get lost again: Vision based
  navigation using streetview images},'' in \emph{Lecture Notes in Computer
  Science (including subseries Lecture Notes in Artificial Intelligence and
  Lecture Notes in Bioinformatics)}.\hskip 1em plus 0.5em minus 0.4em\relax
  Springer, Cham, 2015, vol. 9007, pp. 99--114. [Online]. Available:
  \url{http://link.springer.com/10.1007/978-3-319-16814-2{\_}7}
\BIBentrySTDinterwordspacing

\bibitem{Cowley2011}
\BIBentryALTinterwordspacing
A.~Cowley, C.~J. Taylor, and B.~Southall, ``{Rapid multi-robot exploration with
  topometric maps},'' in \emph{IEEE International Conference on Robotics and
  Automation (ICRA)}.\hskip 1em plus 0.5em minus 0.4em\relax IEEE, may 2011,
  pp. 1044--1049. [Online]. Available:
  \url{http://ieeexplore.ieee.org/document/5980403/}
\BIBentrySTDinterwordspacing

\bibitem{Straub2015}
\BIBentryALTinterwordspacing
J.~Straub, N.~Bhandari, J.~J. Leonard, and J.~W. Fisher, ``{Real-time manhattan
  world rotation estimation in 3D},'' in \emph{IEEE International Conference on
  Intelligent Robots and Systems}, vol. 2015-Decem.\hskip 1em plus 0.5em minus
  0.4em\relax IEEE, sep 2015, pp. 1913--1920. [Online]. Available:
  \url{http://ieeexplore.ieee.org/document/7353628/}
\BIBentrySTDinterwordspacing

\bibitem{Joo2016}
K.~Joo, T.-H. Oh, J.~Kim, and I.~S. Kweon, ``{Globally Optimal Manhattan Frame
  Estimation in Real-Time},'' in \emph{2016 IEEE Conference on Computer Vision
  and Pattern Recognition (CVPR)}, 2016, pp. 1763--1771.

\bibitem{Kschischang2001}
F.~R. Kschischang, B.~J. Frey, and H.~A. Loeliger, ``{Factor graphs and the
  sum-product algorithm},'' \emph{IEEE Transactions on Information Theory},
  2001.

\bibitem{Tibshirani2011}
R.~Tibshirani, ``{Regression shrinkage and selection via the lasso: A
  retrospective},'' \emph{Journal of the Royal Statistical Society. Series B:
  Statistical Methodology}, 2011.

\bibitem{ovc}
``Open vision computer,'' \url{https://github.com/osrf/ovc}.

\end{thebibliography}

\clearpage

\clearpage

\addtolength{\textheight}{-12cm}   





\end{document}